%% file: main.tex
\documentclass[10pt,twocolumn,letterpaper]{article}

\usepackage[pagenumbers]{cvpr} 

\usepackage{graphicx}
\usepackage{amsmath}
\usepackage{amssymb}
\usepackage{booktabs}
\usepackage{array}
\usepackage{multirow}
\usepackage{color}
\usepackage{colortbl}
\usepackage{framed}
\usepackage{bm}
\usepackage{bbm}
\usepackage{xspace}
\usepackage{enumitem}
\usepackage{cite}
\usepackage{xparse}
\usepackage{xcolor}
\usepackage{algorithm}
\usepackage{lipsum}
\usepackage{listings}
\usepackage{url}

\definecolor{Gray}{gray}{0.9}
\definecolor{LightCyan}{rgb}{0.88,0.95,1}

\def \ie {\emph{i.e.}}
\def \eg {\emph{e.g.}}
\def \etal {\emph{et al.}}

\newcommand{\ours}{PAC-S\xspace}
\newcommand{\oursref}{RefPAC-S\xspace}

\newcommand{\tit}[1]{\smallbreak\noindent\textbf{#1.}}
\newcommand{\tinytit}[1]{\noindent\textbf{#1.}}

\usepackage[pagebackref,breaklinks,colorlinks]{hyperref}

\usepackage[capitalize]{cleveref}
\crefname{section}{Sec.}{Secs.}
\Crefname{section}{Section}{Sections}
\Crefname{table}{Table}{Tables}
\crefname{table}{Tab.}{Tabs.}

\def\cvprPaperID{***} 
\def\confName{CVPR}
\def\confYear{2023}

\begin{document}

\title{Positive-Augmented Contrastive Learning for\\Image and Video Captioning Evaluation}

\author{Sara Sarto$^1$ \quad Manuele Barraco$^1$ \quad Marcella Cornia$^1$ \quad Lorenzo Baraldi$^1$ \quad Rita  Cucchiara$^{1,2}$  \\
$^1$University of Modena and Reggio Emilia, Modena, Italy \quad $^2$IIT-CNR, Pisa, Italy\\
{\tt\small \{name.surname\}@unimore.it}
}
\maketitle

\begin{abstract}
The CLIP model has been recently proven to be very effective for a variety of cross-modal tasks, including the evaluation of captions generated from vision-and-language architectures. In this paper, we propose a new recipe for a contrastive-based evaluation metric for image captioning, namely Positive-Augmented Contrastive learning Score (PAC-S), that in a novel way unifies the learning of a contrastive visual-semantic space with the addition of generated images and text on curated data. Experiments spanning several datasets demonstrate that our new metric achieves the highest correlation with human judgments on both images and videos, outperforming existing reference-based metrics like CIDEr and SPICE and reference-free metrics like CLIP-Score. Finally, we test the system-level correlation of the proposed metric when considering popular image captioning approaches, and assess the impact of employing different cross-modal features. Our source code and trained models are publicly available at: {\url{https://github.com/aimagelab/pacscore}}.
\end{abstract}

\section{Introduction}
\label{sec:intro}
\input{sections/01_introduction.tex}

\section{Related Work}
\label{sec:related}
\input{sections/02_related.tex}

\section{Positive-Augmented Contrastive Learning}
\label{sec:method}
\input{sections/03_method.tex}

\section{Experimental Evaluation}
\label{sec:experiments}
\input{sections/04_experiments.tex}

\section{Conclusion}
\label{sec:conclusion}
\input{sections/05_conclusion.tex}

\section*{Acknowledgments}
\small{We thank CINECA for providing computational resources. Work conducted under a research grant co-funded by Leonardo S.p.A. and supported by the projects: PNRR-M4C2 (PE00000013) ``FAIR - Future Artificial Intelligence Research'' funded by the European Commission, ``ELSA - European Lighthouse on Secure and Safe AI'' funded by the EU (GA 101070617), and the PRIN ``CREATIVE: CRoss-modal understanding and gEnerATIon of Visual and tExtual content'' co-funded by the Italian Ministry of University and Research (CUP B87G22000460001).}

{\small
\bibliographystyle{ieee_fullname}
\bibliography{bibliography}
}

\newpage
\newpage
\normalsize
\input{supplementary}

\end{document}

%% file: sections/01_introduction.tex
The task of image captioning, which requires an algorithm to describe visual contents with natural language sentences, has been gaining considerable attention from the research community in the past few years~\cite{xu2015show,vinyals2015show,karpathy2015deep}. As such, the task has witnessed methodological and architectural innovations, ranging from the usage of self-attentive models~\cite{huang2019attention,herdade2019image,cornia2020meshed,pan2020x} to the development of better connections between visual and textual modalities with the addition of objects~\cite{anderson2018bottom,zhang2021vinvl,yang2019auto} and tags~\cite{yao2017boosting,li2020oscar} or the use of more powerful cross-modal features~\cite{barraco2022unreasonable,barraco2022camel,sarto2022retrieval}.

\begin{figure}[t]
    \centering
    \includegraphics[width=\linewidth]{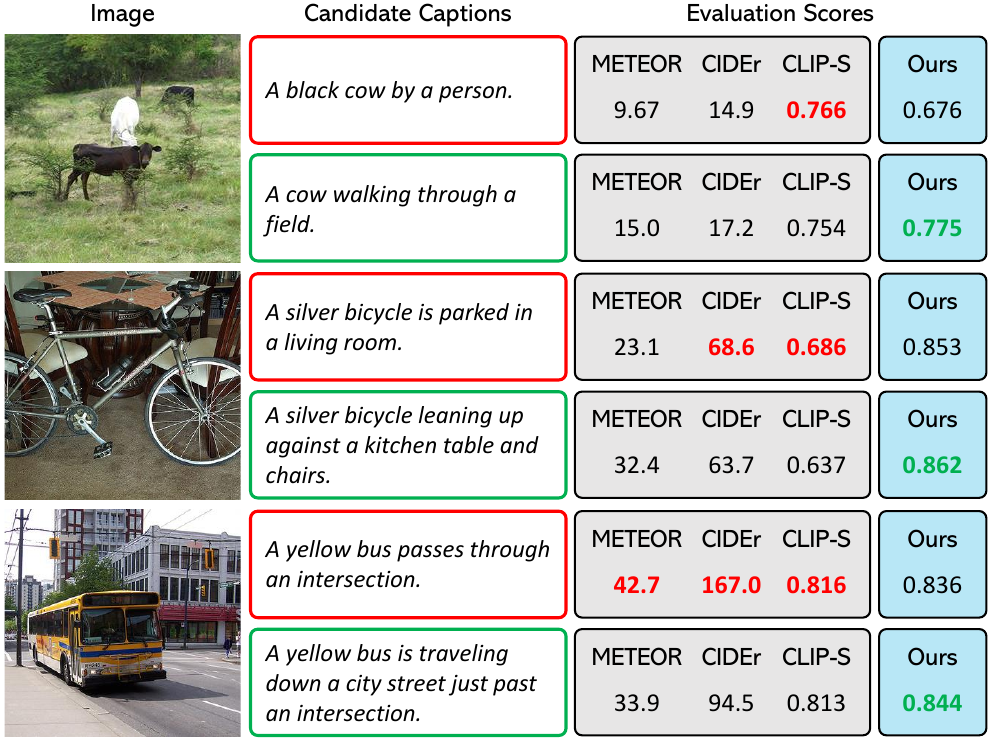}
    \caption{Evaluation scores generated by our proposed metric, \ours, in comparison with existing metrics for captioning. The caption highlighted in green is the one preferred by humans.}
    \label{fig:firstpage}
    \vspace{-.35cm}
\end{figure}

Together with an increase in generation quality, the automatic evaluation of captions has also witnessed a significant effort. While early evaluation scores were based on translation metrics~\cite{papineni2002bleu,lin2004rouge,banerjee2005meteor}, more effective text-based~\cite{vedantam2015cider,spice2016,zhang2019bertscore} and multimodal solutions~\cite{jiang2019tiger,wang2021faier} have been proposed in the last few years. Among these, the usage of cross-modal models in which both visual and textual data can be matched has proven to be a viable strategy that can lead to high-quality metrics~\cite{lee2020vilbertscore,lee2021umic,hessel2021clipscore,kim2022mutual}. Recently, the large-scale CLIP model~\cite{radford2021learning} was tested for image captioning evaluation, resulting in the CLIP-Score~\cite{hessel2021clipscore} which proved to have a significant correlation with human judgment.

While these advancements demonstrate the appropriateness of using contrastive-based embedding spaces for evaluating image captions, large-scale models pre-trained on web-collected data also have limitations, due to the lack in style of captions collected from alt-tags and of the distribution of web-scale images which is not aligned with those on which captioning systems are evaluated. While cleaned data sources, on the contrary, are limited in size, recent advances in both image~\cite{ramesh2022hierarchical,saharia2022photorealistic,rombach2022high,gafni2022make} and text generation~\cite{zhang2021vinvl,wang2022simvlm,li2022blip} have made it possible to synthetically generate data in both modalities, with controlled style and quality.

Following this insight, in this paper we propose a learnable metric that fuses together the advantages of both these scenarios, by leveraging the quality of the pre-training on web-collected data and that of cleaned data, and also regularizing the training by considering additional positive samples hailing from visual and textual generators. Specifically, our proposed metric, PAC-S, is trained via a newly conceived positive-augmented contrastive learning approach, in which pairs of generated images and texts act as additional positives in addition to real images and human-annotated captions taken from a cleaned data source. We demonstrate that the combination of these factors, \ie~the usage of a cleaned data source and the pairing with multimodal generated data, when used to finetune a large-scale contrastive model, results in an embedding space with significantly higher alignment with the human judgment (Fig.~\ref{fig:firstpage}). We apply the resulting metric to evaluate both images and videos, both in reference-based and reference-free settings.

We investigate the quality of the proposed metric by conducting extensive experiments on a variety of image and video datasets, including Flickr8k-Expert and Flickr8k-CF~\cite{hodosh2013framing}, Composite~\cite{aditya2015images}, Pascal-50S, and Abstract-50S~\cite{vedantam2015cider} for the image scenario and the VATEX-EVAL dataset~\cite{shi2022emscore} to evaluate video-caption pairs. Further, we verify its sensitivity to object hallucination on the FOIL~\cite{shekhar2017foil} and ActivityNet-FOIL~\cite{shi2022emscore} datasets and compare the performance of state-of-the-art caption generators with respect to the proposed metric. Our proposal outperforms previous reference-based and reference-free metrics and showcases superior performance with respect to CLIP-Score~\cite{hessel2021clipscore} and the corresponding video-based version (\ie~EMScore~\cite{shi2022emscore}), which also employ a contrastive-based embedding space. Overall, our metric ranks first in terms of correlation with human judgment with respect to all existing image and video captioning metrics.

To sum up, the main contribution of this paper is a novel metric for image and video captioning, based on a positive-augmented training of a multimodal embedding space, which exploits both curated image-caption pairs and additional synthetically generated positives. Extensive experiments on several datasets demonstrate a higher correlation with human judgment and an increased sensitivity to object hallucination.

%% file: sections/02_related.tex
Image and video captioning solutions have been traditionally evaluated using a set of standard evaluation metrics, specifically BLEU~\cite{papineni2002bleu}, METEOR~\cite{banerjee2005meteor}, ROUGE~\cite{lin2004rouge}, CIDEr~\cite{vedantam2015cider}, and SPICE~\cite{spice2016}. Some of them have been originally introduced to evaluate NLP tasks such as machine translation and summarization, while others have been specifically designed for the captioning task. 

Recently, research efforts have been made to introduce additional metrics that can capture different aspects of generated textual sentences, like diversity~\cite{shetty2017speaking,van2018measuring,wang2019describing,wang2020diversity}, robustness of object hallucination~\cite{rohrbach2018object}, uniqueness~\cite{wang2020towards}, and coverage of ground-truth named entities~\cite{cornia2019show,cornia2020smart}. A new trend, instead, is to exploit the capabilities of pre-trained models to compare textual-only~\cite{zhang2019bertscore,yi2020improving} or visual-textual contents~\cite{jiang2019tiger,jiang2019reo,lee2020vilbertscore,wang2021faier,lee2021umic,hessel2021clipscore}. Among them, the BERT score~\cite{zhang2019bertscore} and its improved version~\cite{yi2020improving} use pre-trained BERT embeddings~\cite{devlin2018bert} to represent and compare word tokens in the generated and ground-truth sentences.

\begin{figure*}[t]
    \centering
    \includegraphics[width=0.92\linewidth]{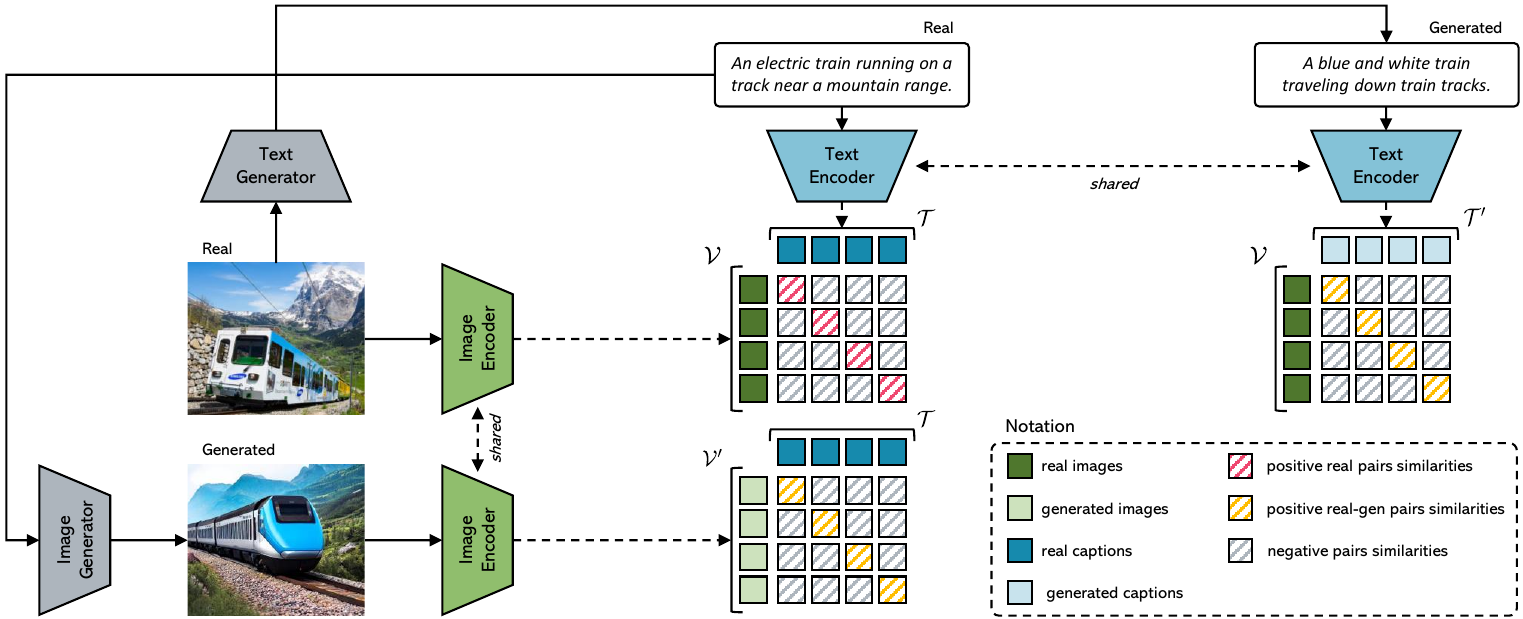}
    \caption{Overview of our positive-augmented contrastive learning approach.}
    \label{fig:model}
    \vspace{-.3cm}
\end{figure*}

In addition to these text-based metrics, other solutions leverage the multimodal nature of vision-and-language models to exploit not only textual information but also the visual content of images and potentially video frames. For example, Jiang~\etal~\cite{jiang2019tiger} introduced the TIGEr metric, which considers the similarities between words and image regions computed according to a cross-modal matching model~\cite{lee2018stacked} trained on COCO~\cite{lin2014microsoft}. Other approaches, instead, exploit the effectiveness of web-scale vision-and-language models such as VilBERT~\cite{lu2019vilbert}, UNITER~\cite{chen2020uniter}, and CLIP~\cite{radford2021learning}, pre-trained on millions or even billions of image-text pairs, to obtain more robust metrics~\cite{lee2020vilbertscore,lee2021umic,hessel2021clipscore,kim2022mutual}. Among them, the recent CLIP-Score~\cite{hessel2021clipscore} is based on a modified cosine similarity between image and candidate caption representations coming from the CLIP model. Recently, Kim~\etal~\cite{kim2022mutual} proposed using CLIP visual-textual features to compute the negative Gaussian cross-mutual information, obtaining a more effective evaluation metric.

While all the aforementioned evaluation metrics have originally been introduced for image captioning, there is only one attempt to evaluate video descriptions through learnable metrics also taking into account the visual content appearing in video frames. In particular, Shi~\etal~\cite{shi2022emscore} presented the EMScore, in its both reference-free and reference-based versions, that computes fine-grained similarities between video frames and words of the candidate caption using CLIP visual-textual embeddings. 

Another related work is that proposed in~\cite{zhu2023imagine} where diffusion models are used to evaluate text-only tasks. Differently from our proposal, the introduced metric exploits similarities between machine-generated images obtained by a visual generator~\cite{rombach2022high} starting from reference and candidate textual items during evaluation.

%% file: sections/03_method.tex
We are interested in devising an image and video captioning metric based on a shared embedding space in which both visual data and text can be projected and compared. To this aim, we start from the dual-encoder architecture popularized by CLIP~\cite{radford2021learning}, which comprises an image encoder~\cite{he2016deep,dosovitskiy2020image} and a text encoder~\cite{vaswani2017attention}. In this architecture, the multimodal interaction is performed in a late fusion fashion, by projecting the output of both encoders to a common dimensionality and then on the $\ell_2$ hypersphere via normalization. The visual and the textual inputs can then be compared via cosine similarity.

Starting from a trained embedding space, an evaluation metric for image captioning can be defined by simply scaling, and eventually thresholding, the similarity computed inside of the embedding itself. For instance, given a visual embedding $v$ and a textual embedding $t$, Hessel~\etal~\cite{hessel2021clipscore} define the evaluation score as
\begin{equation}
    \label{eq:clip_score}
    \text{Score}(t, v) = w \cdot \max (\cos (t, v), 0),
\end{equation}
where $\cos$ indicates the cosine similarity computed inside of the embedding space and $w$ is a scaling factor to enhance numerical readability.

Large-scale contrastive models like CLIP~\cite{radford2021learning} are trained on web-collected image-caption pairs. These provide a large-scale source of supervision for learning scalable low-level and semantic visual and textual features, as testified by their zero-shot classification performance and by their adaptability to different tasks~\cite{ramesh2022hierarchical,barraco2022unreasonable,materzynska2022disentangling,khandelwal2022simple}. Nevertheless, it shall be noted that the textual annotations contained in alt-tags are far from the quality level that a captioning evaluator should look for, and that the distribution of web-scale images might not be properly aligned with those on which image captioning systems are evaluated.

To solve this issue, one might think of learning the metric directly on cleaned data sources. However, recent attempts of learning contrastive-based evaluation metrics on cleaned datasets like COCO~\cite{lin2014microsoft} perform poorly when compared to traditional metrics, potentially because of the lack of training data~\cite{jiang2019tiger}. We, therefore, advocate the usage of synthetic generators of both visual and textual data, which showcase sufficiently high quality levels when generating both images and texts, do lack in terms of style, and are controllable in terms of visual distribution. 

Specifically, given a positive image-text pair $(v, t)$, we augment it by generating a synthetic caption $t'$ from $v$ using an image captioner~\cite{li2022blip}, and a synthetic image $v'$ from $t$ via a diffusion-based text-to-image model~\cite{rombach2022high}, thus building a dataset consisting of tuples of four elements $(v, t, v', t')$. As in Eq.~\ref{eq:clip_score}, we represent $t'$ and $v'$ via their respective text and image embedding. We then train our evaluation model by jointly taking into account contrastive relationships between real and generated matching image-caption pairs (Fig.~\ref{fig:model}). To lower the computational requirements, we start with pre-trained CLIP visual and textual encoders and only train the projection toward the embedding space. 

Formally, given a batch of $N$ real images $\mathcal{V} = \left[ v_1, v_2, ..., v_N \right]$ and their corresponding captions $\mathcal{T} = \left[ t_1, t_2, ..., t_N \right]$, generated images $\mathcal{V}' = \left[ v_1', v_2', ..., v_N' \right]$ and generated texts $\mathcal{T}' = \left[ t_1', t_2', ..., t_N' \right]$, we define multiple $N \times N$ matrices containing pairwise cosine similarities between the different inputs. We then adopt a symmetric InfoNCE loss~\cite{oord2018representation} which aims at maximizing the cosine similarity between the $N$ matching pairs and minimize those of the $N^2 - N$ non-matching pairs. The loss which compares real images $\mathcal{V}$ with respect to real texts $\mathcal{T}$ can be defined, for instance, as
\begin{gather}
    L_{\mathcal{V}, \mathcal{T}} = -\frac{1}{N} \sum_{i=1}^N \log \frac{\exp(\cos(v_i, t_i) / \tau)}{\sum_{j=1}^N \exp(\cos(v_i, t_j) / \tau)} + \nonumber \\ 
    -\frac{1}{N} \sum_{i=1}^N \log \frac{\exp(\cos(v_i, t_i) / \tau)}{\sum_{j=1}^N \exp(\cos(v_j, t_i) / \tau)},
\end{gather}
where $\tau$ is a temperature parameter. In addition to a loss term between real images and real texts, $L_{\mathcal{V}, \mathcal{T}}$, we also add symmetrical loss terms between cross-modal generated and real pairs, \ie~between generated images and human-annotated texts, and between original images and generated texts. In this way, generated items act as additional positive samples for the real matching pairs, thus adding a supervisory signal without paying the cost of the noisy data on which contrastive-based features extractors like CLIP are learned. In summary, the final loss is a weighted combination of the three loss terms, \ie 
\begin{equation}
    L = L_{\mathcal{V}, \mathcal{T}} + \lambda_v L_{\mathcal{V}', \mathcal{T}} + \lambda_t  L_{\mathcal{V}, \mathcal{T}'},
\end{equation}
where $L_{\mathcal{V}', \mathcal{T}}$ is the loss between generated images and real texts and $L_{\mathcal{V}, \mathcal{T}'}$ its counterpart between generated texts and real images.

\subsection{Captioning evaluation score for images}
After training with positive-augmented contrastive learning, we employ two evaluation scores for evaluating images in both a reference-free and a reference-based setting. Specifically, we employ Eq.~\ref{eq:clip_score} with $w=2$\footnote{To stretch the range of the score distribution in $\left[0, 1\right]$.} as our reference-free score. Then, we follow the approach proposed in~\cite{hessel2021clipscore} to include reference ground-truth captions in the evaluation process. Specifically, we compute the representation of each reference caption using the textual encoder. Then, we compute the harmonic mean between the reference-free score (Eq.~\ref{eq:clip_score}) and the maximum cosine similarity between the candidate caption and all reference captions. Formally, given a set of reference captions $R = \{ r_1, r_2, ..., r_m\}$, the score is computed as
\begin{gather}
    \text{Ref-Score}(t, v, R) = \text{H-Mean}(\text{Score}(t, v), \nonumber \\ 
    \max(0, \max_{r \in R} \cos (t, r))),
\end{gather}
where $\text{Score}(\cdot)$ indicates the reference-free evaluation score as reported by our positive-augmented embedding space, and $\text{H-Mean}(\cdot)$ indicates the harmonic mean.

\subsection{Captioning evaluation score for videos}
To test the proposed positive-augmented strategy for evaluating video captions, we extend the above defined metric following the approach of~\cite{shi2022emscore}. In this case, matching scores are computed at two granularity levels, \ie~a coarse-grained level in which the global representation of the candidate caption is compared with the global representation of the video, and a fine-grained level in which the embeddings of single words are compared to those of single frames.

Specifically, we use the positive-augmented CLIP visual encoder to extract the embeddings of single frames and average-pool them to get the representation of the entire video. Similarly, we employ the corresponding textual encoder to get single tokens and whole caption embeddings. The fine-grained score is then computed by taking the F1-score of pairwise word-frame similarities and TF-IDF~\cite{robertson2004understanding} weighting, and the coarse-grained score is computed as the similarity between the global video and caption representations. Given a source video $V$ and a candidate caption $c$, the overall score is defined as
\begin{equation}
    \label{eq:emscore}
    \text{Score}(t, V) = \frac{\text{Score}(t, V)_c + \text{Score}(t, V)_f}{2},
\end{equation}
where $\text{Score}_c$ represents the coarse-grained embedding matching and $\text{Score}_f$ stands for the fine-grained similarity.
Finally, to include a set of reference captions $R$, we follow the reference version of the aforementioned approach:
\begin{equation}
    \label{eq:ref_emscore}
    \text{Ref-Score}(t, V, r) = \frac{\text{Score}(t, V) + \max_{r \in R}\text{Score}(t, r)}{2},
\end{equation}
where $\text{Score}(t, r)$ is computed as defined in Eq.~\ref{eq:emscore} by using the word-level embeddings of the reference caption.

\begin{figure}[t]
\centering
\includegraphics[width=\linewidth]{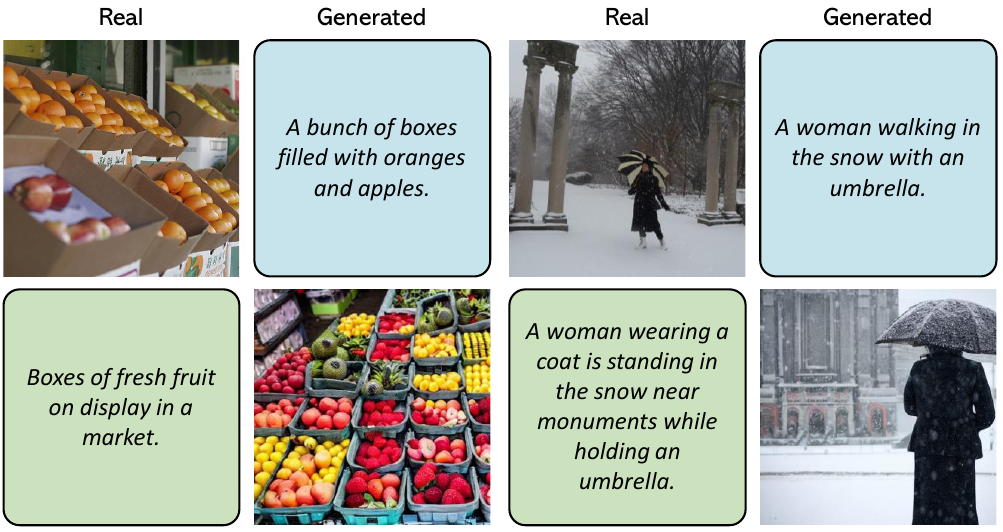}
\caption{Sample real and generated image-text data used for positive-augmented contrastive learning.}
\label{fig:dataset}
\vspace{-.35cm}
\end{figure}

%% file: sections/04_experiments.tex
\subsection{Implementation details}\label{sec:details}
\tinytit{Architecture and training details} In continuity with existing literature~\cite{hessel2021clipscore,kim2022mutual,shi2022emscore}, we use CLIP ViT-B/32~\cite{radford2021learning} as backbone to encode images (or video frames) and textual sentences. We finetune the visual and textual final projections of the model using the approach described in Sec.~\ref{sec:method} on the COCO dataset~\cite{lin2014microsoft}, which contains more than 120k images annotated with five captions. In particular, we employ the splits introduced by Karpathy~\etal~\cite{karpathy2015deep}, where 5,000 images are used for validation, 5,000 images are used for test and the rest for training. 
During finetuning, we use AdamW~\cite{loshchilov2019decoupled} as optimizer with a learning rate equal to 0.0001 and a batch size of 256. The $\lambda_v$ and $\lambda_t$ values are selected with a grid search, choosing the combination that provides the best average across datasets. Specifically, we set $\lambda_v$ to 0.05 and $\lambda_t$ to 0.1, and stop the training stage when the validation loss stops decreasing for 1,500 iterations.

\begin{table}[t]
\small
\centering
\setlength{\tabcolsep}{.3em}
\resizebox{0.95\linewidth}{!}{
\begin{tabular}{lc cc c cc}
\toprule
& & \multicolumn{2}{c}{\textbf{Flickr8k-Expert}} & & \multicolumn{2}{c}{\textbf{Flickr8k-CF}} \\
\cmidrule{3-4} \cmidrule{6-7}
& & Kendall $\tau_b$ & Kendall $\tau_c$  & & Kendall $\tau_b$ & Kendall $\tau_c$ \\
\midrule
BLEU-1~\cite{papineni2002bleu} & & 32.2 & 32.3 & & 17.9 & 9.3 \\
BLEU-4~\cite{papineni2002bleu} & & 30.6 & 30.8 & & 16.9 & 8.7 \\
ROUGE~\cite{lin2004rouge} & & 32.1 & 32.3 & & 19.9 & 10.3 \\
METEOR~\cite{banerjee2005meteor} & & 41.5 & 41.8 & & 22.2 & 11.5 \\
CIDEr~\cite{vedantam2015cider} & & 43.6 & 43.9 & & 24.6 & 12.7 \\
SPICE~\cite{spice2016} & & 51.7 & 44.9 & & 24.4 & 12.0 \\
\midrule
BERT-S~\cite{zhang2019bertscore} & & - & 39.2 & & 22.8 & - \\
LEIC~\cite{cui2018learning} & & 46.6 & - & & 29.5 & - \\
BERT-S++~\cite{yi2020improving} & & - & 46.7 & & - & - \\
UMIC~\cite{lee2021umic} & & - & 46.8 & & - & - \\
TIGEr~\cite{jiang2019tiger} & & - & 49.3 & & - & - \\
ViLBERTScore~\cite{lee2020vilbertscore} & & - & 50.1 & & - & - \\
MID~\cite{kim2022mutual} & & - & 54.9  & & 37.3 & - \\
\midrule
CLIP-S~\cite{hessel2021clipscore} & & 51.1 & 51.2 & & 34.4 & 17.7 \\
\rowcolor{LightCyan}
& & \underline{53.9} & \underline{54.3} & & \underline{36.0} & \underline{18.6}  \\
\rowcolor{LightCyan}
\multirow{-2}{*}{\textbf{\ours}} & & (\textcolor{blue}{+2.8}) & (\textcolor{blue}{+3.1}) & & (\textcolor{blue}{+1.6}) & (\textcolor{blue}{+0.9}) \\
\midrule
RefCLIP-S~\cite{hessel2021clipscore} & & 52.6 & 53.0 & & 36.4 & 18.8 \\
\rowcolor{LightCyan}
& & \underline{\textbf{55.5}} & \underline{\textbf{55.9}} & & \underline{\textbf{37.6}} & \underline{\textbf{19.5}}  \\
\rowcolor{LightCyan}
\multirow{-2}{*}{\textbf{\oursref}} & & (\textcolor{blue}{+2.9}) & (\textcolor{blue}{+2.9}) & & (\textcolor{blue}{+1.2}) & (\textcolor{blue}{+0.7}) \\
\bottomrule
\end{tabular}
 }
 \vspace{-0.1cm}
\caption{Human judgment correlation scores on Flickr8k-Expert and Flickr8k-CF~\cite{hodosh2013framing}. The overall best scores are in bold.}
\label{tab:flickr}
\vspace{-0.35cm}
\end{table}

\tit{Positive image-text generation} To augment the training set with new positive examples, we use Stable Diffusion\footnote{\url{https://github.com/CompVis/stable-diffusion}}~\cite{rombach2022high} for generating new visual data and the BLIP model~\cite{li2022blip} for generating new textual descriptions. Specifically, to generate images, we employ the model pre-trained on the English image-text pairs of the LAION-5B dataset~\cite{schuhmann2022laion} and finetuned at a resolution equal to $512\times512$ on the LAION-Aesthetics subset\footnote{\url{https://laion.ai/blog/laion-aesthetics/}}, which has been filtered with aesthetic requirements. 
During generation, we employ the safety checker module to reduce the probability of explicit images and disable the invisible watermarking of the outputs to avoid easy identification of the images as machine-generated. To generate text, instead, we use the ViT-L/14 version\footnote{\url{https://github.com/salesforce/BLIP}} of the BLIP model pre-trained on 129M image-text pairs and finetuned on the COCO dataset. After this generation phase, we get a new version of the COCO dataset in which each image is additionally associated with a machine-generated caption and each human-annotated caption is instead associated with a newly generated image. Sample image-text data employed for finetuning are shown in Fig.~\ref{fig:dataset}. 

\subsection{Correlation with human judgment}
To evaluate the correlation of the proposed metric with human ratings, we conduct experiments on both image and video captioning datasets. Specifically, we employ the Flickr8k-Expert, Flickr8k-CF, and Composite datasets~\cite{hodosh2013framing,aditya2015images} for the image setting and the VATEX-EVAL dataset~\cite{shi2022emscore} to evaluate video-caption pairs.

\tit{Image captioning results} We first evaluate our solution on the Flickr8k-Expert and Flickr8k-CF datasets~\cite{hodosh2013framing} which include image-caption pairs with corresponding human ratings. In particular, Flickr8k-Expert contains 17k expert annotations for visual-textual pairs, with a total of 5,664 different images. The pairs are evaluated with a score from 1 to 4, where 1 indicates that the caption does not correlate with the image and 4 indicates that the caption describes the corresponding image without errors. Flickr8k-CF, instead, is composed of 145k binary quality judgments, collected from CrowdFlower, for 48k image-caption pairs (with 1,000 unique images). Each pair is annotated with at least three binary scores, where ``yes'' indicates that the caption correlates with the image. To measure the correlation with human judgment, we compute the mean proportion of ``yes'' annotations as the score for each pair. 

\begin{table}[t]
\small
\centering
\setlength{\tabcolsep}{.55em}
\resizebox{.68\linewidth}{!}{
\begin{tabular}{lc cc}
\toprule
& & \multicolumn{2}{c}{\textbf{Composite}} \\
\cmidrule{3-4}
& & Kendall $\tau_b$ & Kendall $\tau_c$ \\
\midrule
BLEU-1~\cite{papineni2002bleu} & & 29.0 & 31.3 \\
BLEU-4~\cite{papineni2002bleu} & & 28.3 & 30.6 \\
ROUGE~\cite{lin2004rouge} & & 30.0 & 32.4 \\
METEOR~\cite{banerjee2005meteor} & & 36.0 & 38.9 \\
CIDEr~\cite{vedantam2015cider} & & 34.9 & 37.7 \\
SPICE~\cite{spice2016} & & 38.8 & 40.3 \\
\midrule
BERT-S~\cite{zhang2019bertscore} & & - & 30.1 \\
BERT-S++~\cite{yi2020improving} & & - & 44.9 \\
TIGEr~\cite{jiang2019tiger} & & - & 45.4 \\
ViLBERTScore~\cite{lee2020vilbertscore} & & - & 52.4 \\
FAIEr~\cite{wang2021faier} & & - & 51.4 \\
\midrule
CLIP-S~\cite{hessel2021clipscore} & & 49.8 & 53.8 \\
\rowcolor{LightCyan}
 & & \underline{51.5} & \underline{55.7} \\
\rowcolor{LightCyan}
\multirow{-2}{*}{\textbf{\ours}} & & (\textcolor{blue}{+1.7}) & (\textcolor{blue}{+1.9}) \\
\midrule
RefCLIP-S~\cite{hessel2021clipscore} & & 51.2 & 55.4 \\
\rowcolor{LightCyan}
 & & \underline{\textbf{53.0}} & \underline{\textbf{57.3}} \\
 \rowcolor{LightCyan}
\multirow{-2}{*}{\textbf{\oursref}} & & (\textcolor{blue}{+1.8}) & (\textcolor{blue}{+1.9}) \\
\bottomrule
\end{tabular}
}
\vspace{-0.1cm}
\caption{Human judgment correlation scores on the Composite dataset~\cite{aditya2015images}. The overall best scores are in bold.}
\label{tab:composite}
\vspace{-0.35cm}
\end{table}

\begin{table*}[t]
\small
\centering
\begin{minipage}{0.6\linewidth}
\setlength{\tabcolsep}{.3em}
\resizebox{\linewidth}{!}{
\begin{tabular}{lc cc c cc c cc}
\toprule
& & \multicolumn{2}{c}{\textbf{No Ref}} & & \multicolumn{2}{c}{\textbf{1 Ref}} & & \multicolumn{2}{c}{\textbf{9 Refs}} \\
\cmidrule{3-4} \cmidrule{6-7} \cmidrule{9-10}
& & Kendall $\tau_b$ & Spearman $\rho$ & & Kendall $\tau_b$ & Spearman $\rho$ & & Kendall $\tau_b$ & Spearman $\rho$ \\
\midrule
BLEU-1~\cite{papineni2002bleu} & & - & - & & 12.2 & 15.9 & & 28.9 & 37.0 \\
BLEU-4~\cite{papineni2002bleu} & & - & - & & 12.6 & 16.4 & & 22.4 & 29.5 \\
ROUGE~\cite{lin2004rouge} & & - & - & & 12.5 & 16.3 & & 23.8 & 30.9 \\
METEOR~\cite{banerjee2005meteor} & & - & - & & 16.4 & 21.5 & & 27.6 & 35.7 \\
CIDEr~\cite{vedantam2015cider} & & - & - & &  17.3 & 22.6 & & 27.8 & 36.1 \\
\midrule
BERT-S~\cite{zhang2019bertscore} & & - & - & & 18.2 & 23.7 & & 29.3 & 37.8 \\
BERT-S++~\cite{yi2020improving} & & - & - & & 15.2 & 19.8 & & 24.4 & 31.7 \\
\midrule
EMScore~\cite{shi2022emscore} & & 23.2 & 30.3 & & 28.6 & 37.1 & & 36.8 & 47.2 \\
\rowcolor{LightCyan}
& & \underline{\textbf{25.1}} & \underline{\textbf{32.6}} & & \underline{\textbf{31.4}} & \underline{\textbf{40.5}} & & \underline{\textbf{38.1}} & \underline{\textbf{48.8}} \\
\rowcolor{LightCyan}
\multirow{-2}{*}{\textbf{\ours~/ \oursref}} & & (\textcolor{blue}{+1.9}) & (\textcolor{blue}{+2.3}) & & (\textcolor{blue}{+2.8}) & (\textcolor{blue}{+3.4}) & & (\textcolor{blue}{+1.3}) & (\textcolor{blue}{+1.6})  \\
\bottomrule
\end{tabular}
}
\end{minipage}
\hfill
\begin{minipage}{0.37\linewidth}
\includegraphics[width=0.95\textwidth]{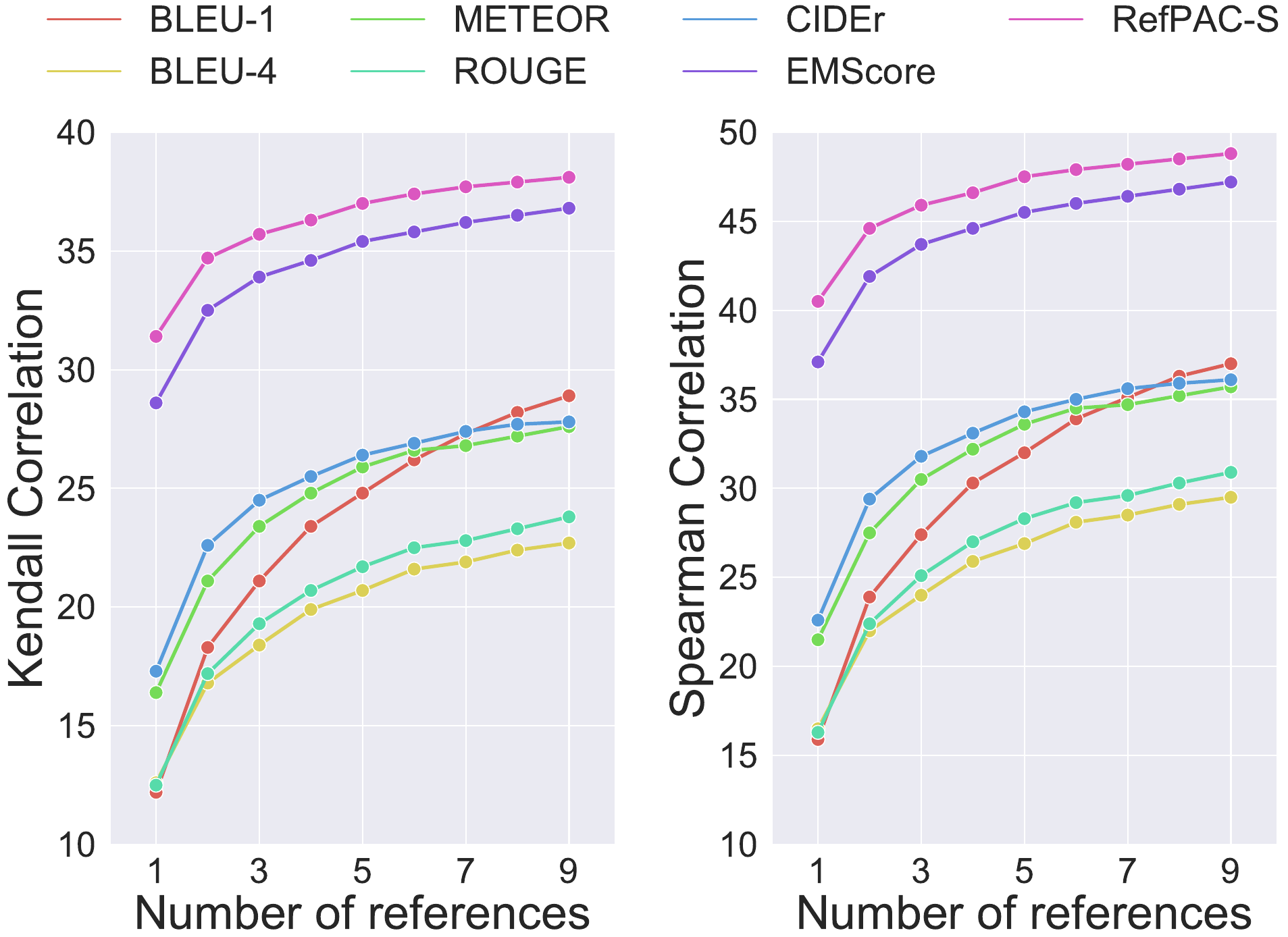}
\end{minipage}
\vspace{-0.1cm}
\caption{Human judgment correlation scores on the VATEX-EVAL dataset~\cite{shi2022emscore}. The overall best scores are in bold. On the right, we show Kendall $\tau_b$ correlation score at varying of the number of reference captions.}
\label{tab:vatex}
\vspace{-0.3cm}
\end{table*}

Following previous works~\cite{zhang2019bertscore,lee2020vilbertscore,lee2021umic,hessel2021clipscore}, we compute Kendall correlation scores in both $\tau_b$ and $\tau_c$ versions. Results are reported in Table~\ref{tab:flickr} comparing the proposed \ours metric with respect to both standard captioning evaluation scores (\ie~BLEU~\cite{papineni2002bleu}, ROUGE~\cite{lin2004rouge}, METEOR~\cite{banerjee2005meteor}, CIDEr~\cite{vedantam2015cider}, and SPICE~\cite{spice2016}) and more recent solutions that either exploit text-only or cross-modal learned embeddings, such as BERT-S~\cite{zhang2019bertscore}, BERT-S++~\cite{yi2020improving}, LEIC~\cite{cui2018learning}, TIGEr~\cite{jiang2019tiger}, UMIC~\cite{lee2021umic}, VilBERTScore~\cite{lee2020vilbertscore}, MID~\cite{kim2022mutual}, and CLIP-S~\cite{hessel2021clipscore}. While CLIP-S is reported in both reference-free and reference-based versions, all other metrics require reference captions. The only exception is the MID score which is positioned between a reference-free and a reference-based metric since it utilizes the mean and covariance of the correct captions. 

From the results, it can be seen that the proposed score achieves the best correlation with human judgment on both considered datasets, demonstrating its effectiveness compared to previously proposed metrics. In particular, when comparing our score with CLIP-S and RefCLIP-S, we can notice an improvement in terms of Kendall $\tau_b$ of 2.8 and 2.9 points on Flickr8k-Expert, and 1.6 and 1.2 points on Flickr8k-CF, respectively. Similar improvements can be also observed in terms of Kendall $\tau_c$ correlation score. It is also important to note that the reference-free version of \ours overcomes by a large margin the correlation scores achieved by traditional reference-based metrics such as CIDEr and SPICE (\eg~+10.3/10.4 points with respect to the CIDEr metric on Flickr8k-Expert).

We also conduct experiments on the Composite dataset~\cite{aditya2015images} which contains 12k human judgments for image-caption pairs taken from COCO~\cite{lin2014microsoft} (2,007 images), Flickr8k~\cite{hodosh2013framing} (997 images), and Flickr30k~\cite{young2014image} (991 images). Each image-caption pair is evaluated with a score, given by humans, between 1 and 5 to estimate the correspondence of the caption with the associated image. Experimental results are shown in Table~\ref{tab:composite}, again in terms of Kendall $\tau_b$ and Kendall $\tau_c$ correlation scores. Also in this case, our metric achieves a better correlation with human ratings than that obtained by both traditional and more recent evaluation scores, confirming its effectiveness even when compared to CLIP-S and RefCLIP-S.

\tit{Video captioning results} To evaluate the correlation with humans in the context of video-caption pairs, we consider the VATEX-EVAL dataset~\cite{shi2022emscore} which includes 3,000 videos from the VATEX~\cite{wang2019vatex} validation set, each of them associated with six captions of mixed quality. Each video-caption pair has been evaluated by three human annotators with a score from 1 (to denote inconsistency between the video and the caption) to 5 (to denote consistency). Overall, the dataset contains 54k human ratings for 18k video-caption pairs. Following recent literature~\cite{shi2022emscore}, we compute Kendall $\tau_b$ and Spearman $\rho$ rank correlation coefficients, considering a different number of reference sentences when measuring correlation (\ie~zero, one, or nine). Correlation scores are reported in Table~\ref{tab:vatex} in comparison with standard evaluation metrics, BERT-S, BERT-S++, and the only video-specific captioning metric existing in literature, \ie~EMScore. On the right, we also report the correlation scores at varying the number of reference captions. It can be seen that PAC-S achieves the best correlation scores in all settings, improving EMScore of 2.3, 3.4, and 1.6 Spearman $\rho$ points respectively with no references, one reference, and nine reference sentences. These results further confirm the appropriateness of our positive-augmented contrastive learning strategy to improve captioning evaluation also when considering videos instead of static images.

\begin{table}[t]
\small
\centering
\setlength{\tabcolsep}{.55em}
\resizebox{0.95\linewidth}{!}{
\begin{tabular}{lc cccc c c}
\toprule
 & & HC & HI & HM & MM & & Mean \\
\midrule
length & & 51.7 & 52.3 & 63.6 & 49.6 & & 54.3 \\
BLEU-1~\cite{papineni2002bleu} & & 64.6 & 95.2 & 91.2 & 60.7 & & 77.9 \\
BLEU-4~\cite{papineni2002bleu} & & 60.3 & 93.1 & 85.7 & 57.0 & & 74.0 \\
ROUGE~\cite{lin2004rouge} & & 63.9 & 95.0 & 92.3 & 60.9 & & 78.0 \\
METEOR~\cite{banerjee2005meteor} & & 66.0 & 97.7 & 94.0 & 66.6 & & 81.1 \\
CIDEr~\cite{vedantam2015cider} & & 66.5 & 97.9 & 90.7 & 65.2 & & 80.1 \\
\midrule
BERT-S$^\dagger$~\cite{zhang2019bertscore} & & 65.4 & 96.2 & 93.3 & 61.4 & & 79.1 \\
BERT-S++$^\dagger$~\cite{yi2020improving} & & 65.4 & 98.1 & 96.4 & 60.3 & & 80.1 \\
TIGEr$^\dagger$~\cite{jiang2019tiger} & & 56.0 & 99.8 & 92.8 & 74.2 & & 80.7 \\
ViLBERTScore$^\dagger$~\cite{lee2020vilbertscore} & & 49.9 & 99.6 & 93.1 & 75.8 & & 79.6\\
FAIEr$^\dagger$~\cite{wang2021faier} & & 59.7 & {\textbf{99.9}} & 92.7 & 73.4 & & 81.4 \\
MID$^\dagger$~\cite{kim2022mutual} & & 67.0 & 99.7 & \textbf{97.4} & \textbf{76.8} & & \textbf{85.2} \\
\midrule
CLIP-S~\cite{hessel2021clipscore} & & 55.9 & \underline{99.3} & 96.5 & 72.0 & & 80.9 \\
\rowcolor{LightCyan}
 & & \underline{60.6} & \underline{99.3} & \underline{96.9} & \underline{72.9} & & \underline{82.4} \\
\rowcolor{LightCyan}
\multirow{-2}{*}{\textbf{\ours}} & & (\textcolor{blue}{+4.7}) & (+0.0) & (\textcolor{blue}{+0.4}) & (\textcolor{blue}{+0.9}) & & (\textcolor{blue}{+1.5}) \\
\midrule
RefCLIP-S~\cite{hessel2021clipscore} & & 64.9 & 99.5 & 95.5 & 73.3 & & 83.3 \\
\rowcolor{LightCyan}
& & \underline{\textbf{67.7}} & \underline{99.6} & \underline{96.0} & \underline{75.6} & & \underline{84.7} \\
\rowcolor{LightCyan}
\multirow{-2}{*}{\textbf{\textbf{\oursref}}} & & (\textcolor{blue}{+2.8}) & (\textcolor{blue} {+0.1}) & (\textcolor{blue}{+0.5}) & (\textcolor{blue}{+2.3}) & & (\textcolor{blue}{+1.4}) \\
\bottomrule
\end{tabular}
}
\caption{Accuracy results on the Pascal-50S dataset~\cite{vedantam2015cider} obtained by averaging the scores over five random draws of reference captions (except for reference-free metrics). The $\dagger$ marker indicates scores reported in previous works, which may differ in terms of selected reference captions. We refer to the text for the definition of HC, HI, HM, and MM. The overall best scores are in bold.}
\label{tab:pascal}
\vspace{-0.35cm}
\end{table}

\subsection{Caption pairwise ranking}
We assess the effectiveness of the proposed metric on the Pascal-50S dataset~\cite{vedantam2015cider}, which reports pairwise preference judgments between two captions. Specifically, the dataset comprises 4k sentence pairs, each of them associated with an image from the UIUC Pascal sentence dataset~\cite{rashtchian2010collecting}. 
For each pair, 48 human judgments have been collected, in which each evaluation expresses which sentence best describes the given image. Sentence pairs are divided into four different categories: two human-written and correct captions (HC), two human-written captions where one is correct and the other is wrong (HI), two correct captions but one written by humans and the other machine-generated (HM), two machine-generated and correct captions (MM).

\begin{table}[t]
\small
\centering
\setlength{\tabcolsep}{.55em}
\resizebox{0.8\linewidth}{!}{
\begin{tabular}{lcc cc lcc}
\toprule
& & Mean & & & & & Mean\\
\midrule
CLIP-S~\cite{hessel2021clipscore} & & 68.2 & & & RefCLIP-S~\cite{hessel2021clipscore} & & 75.8 \\
\rowcolor{LightCyan}
 & & \underline{69.7} & & & & & \underline{\textbf{76.9}} \\
\rowcolor{LightCyan}
\multirow{-2}{*}{\textbf{\ours}} & & (\textcolor{blue}{+1.5}) & & &  \multirow{-2}{*}{\textbf{\oursref}} & & (\textcolor{blue}{+1.1})  \\
\bottomrule
\end{tabular}
}
 \vspace{-0.1cm}
\caption{Accuracy results on the Abstract-50S dataset~\cite{vedantam2015cider}.}
\label{tab:abstract}
\vspace{-0.35cm}
\end{table}

In this setting, instead of computing correlation scores, we compute accuracy by considering for each pair the caption preferred by the majority of human ratings as correct (where ties are broken randomly) and measuring how often the evaluation metric assigns a higher score to the selected caption. For each evaluation, we randomly sample five reference captions (among the 48 provided by the dataset) and average the results over five different draws. Accuracy values are reported in Table~\ref{tab:pascal} in comparison with previously proposed metrics. Similarly to previous works, we also include the results of a length-based baseline in which the longer caption is always considered the better one. From the results, we can observe that \ours and \oursref respectively perform better than CLIP-S and RefCLIP-S in almost all categories, with an increase of 1.5 points in terms of averaged accuracy. Also, our results are generally higher than those of the other metrics, with the only exception of the MID score which achieves slightly better accuracy. However, our results are not directly comparable to the ones reported in previous works (as, for example, FAIEr and MID), given the random selection of ground-truth sentences used to compute reference-based metrics.

As a further analysis, we evaluate the results on the Abstract-50S dataset~\cite{vedantam2015cider} which contains clip-art images from~\cite{zitnick2013bringing} associated with 48 human-annotated reference sentences. Similar to Pascal-50S, each image is associated with a pair of candidate captions and 48 human judgments, collected asking to select which candidate caption is most similar to a given reference sentence. Overall, the dataset is composed of 400 candidate caption pairs, of which 200 describe the corresponding image (\ie~both captions are correct) and 200 instead contain one correct caption and one caption of another image. Again, we compute accuracy scores by considering the most preferred caption as correct, averaging the results over five random draws of reference sentences. Table~\ref{tab:abstract} shows the results of our score in comparison with CLIP-S. In both reference-free and reference-based versions, \ours achieves better accuracy scores than CLIP-S, demonstrating its effectiveness also in this challenging setting of non-photographic images.

\begin{table}[t]
\small
\centering
\setlength{\tabcolsep}{.4em}
\resizebox{.94\linewidth}{!}{
\begin{tabular}{lc cc c c}
\toprule
 & & \multicolumn{2}{c}{\textbf{FOIL}} & & \textbf{ActivityNet-FOIL} \\
 \cmidrule{3-4} \cmidrule{6-6}
 & & Acc. (1 Ref) & Acc. (4 Refs) & & Accuracy \\
\midrule
BLEU-1~\cite{papineni2002bleu} & & 65.7 & 85.4 & & 60.1 \\
BLEU-4~\cite{papineni2002bleu} & & 66.2 & 87.0 & & 66.1 \\
ROUGE~\cite{lin2004rouge} & & 54.6 & 70.4 & & 56.7 \\
METEOR~\cite{banerjee2005meteor} & & 70.1 & 82.0 & & 72.9 \\
CIDEr~\cite{vedantam2015cider} & & 85.7 & 94.1 & & 77.9 \\
MID~\cite{kim2022mutual} & & 90.5 & 90.5 & & - \\
\midrule
CLIP-S~\cite{hessel2021clipscore} & & 87.2 & 87.2 & & - \\
EMScore~\cite{shi2022emscore} & & - & - & & 89.5 \\
\rowcolor{LightCyan}
& & \underline{89.9} & \underline{89.9} & & \underline{90.1} \\
\rowcolor{LightCyan}
\multirow{-2}{*}{\textbf{\ours}} & & (\textcolor{blue}{+2.7}) & (\textcolor{blue}{+2.7}) & & (\textcolor{blue}{+0.6}) \\
\midrule
RefCLIP-S~\cite{hessel2021clipscore} & & 91.0 & 92.6 & & - \\
EMScoreRef~\cite{shi2022emscore} & & - & - & & 92.4 \\
\rowcolor{LightCyan}
& & \underline{\textbf{93.7}} & \underline{\textbf{94.9}} & & \underline{\textbf{93.5}} \\
\rowcolor{LightCyan}
\multirow{-2}{*}{\textbf{\oursref}} & & (\textcolor{blue}{+2.7}) & (\textcolor{blue}{+2.3}) & & (\textcolor{blue}{+1.1}) \\
\bottomrule
\end{tabular}
}
 \vspace{-0.1cm}
\caption{Accuracy results on the FOIL~\cite{shekhar2017foil} and ActivityNet-FOIL~\cite{shi2022emscore} hallucination detection datasets. The overall best scores are in bold.}
\label{tab:foil}
\vspace{-0.2cm}
\end{table}

\begin{table}[t]
\small
\centering
\setlength{\tabcolsep}{.32em}
\resizebox{\linewidth}{!}{
\begin{tabular}{lc ccc ccc}
\toprule
& & B-4 & M & C & CLIP-S & \textbf{\ours} & \textbf{\oursref} \\
\midrule
Show and Tell~\cite{vinyals2015show} & & 31.4 & 25.0 & 97.2 & 0.572 & \cellcolor[rgb]{0.88,0.95,1} 0.772 & \cellcolor[rgb]{0.88,0.95,1} 0.826 \\
Show, Attend and Tell~\cite{xu2015show} & & 33.4 & 26.2 & 104.6 & 0.582 & \cellcolor[rgb]{0.88,0.95,1} 0.785 & \cellcolor[rgb]{0.88,0.95,1} 0.837 \\
Up-Down~\cite{anderson2018bottom} & &  36.7 & 27.9 & 122.7 & 0.592 & \cellcolor[rgb]{0.88,0.95,1} 0.794 & \cellcolor[rgb]{0.88,0.95,1} 0.847 \\
SGAE~\cite{yang2019auto} & & 39.0 & 28.4 & 129.1 & 0.600 & \cellcolor[rgb]{0.88,0.95,1} 0.803 & \cellcolor[rgb]{0.88,0.95,1} 0.854 \\
AoANet~\cite{huang2019attention} & & 38.9 & 29.2 & 129.8 &  0.602 & \cellcolor[rgb]{0.88,0.95,1} 0.805 & \cellcolor[rgb]{0.88,0.95,1} 0.856 \\
$\mathcal{M}^2$ Transformer~\cite{cornia2020meshed} & & 39.1 & 29.2 & 131.2 & 0.605 & \cellcolor[rgb]{0.88,0.95,1} 0.806 & \cellcolor[rgb]{0.88,0.95,1} 0.854 \\
X-Transformer~\cite{pan2020x} & & 39.7 & 29.5 & 132.8 & 0.610 & \cellcolor[rgb]{0.88,0.95,1} 0.812 & \cellcolor[rgb]{0.88,0.95,1} 0.859 \\
VinVL~\cite{zhang2021vinvl} & & \textbf{41.0} & \textbf{31.1} & \textbf{140.9} & \textbf{0.627} & \cellcolor[rgb]{0.88,0.95,1} \textbf{0.821} & \cellcolor[rgb]{0.88,0.95,1} \textbf{0.869} \\
\midrule
\textit{Humans} & & - & \textit{24.1} & \textit{87.6} & \textit{0.626} & \cellcolor[rgb]{0.88,0.95,1} \textit{0.818} & \cellcolor[rgb]{0.88,0.95,1} \textit{0.857} \\
\bottomrule
\end{tabular}
}
 \vspace{-0.1cm}
\caption{Evaluation scores of state-of-the-art captioning models on COCO test set~\cite{lin2014microsoft}.}
\label{tab:sota}
\vspace{-0.35cm}
\end{table}

\begin{table*}[t]
\small
\centering
\setlength{\tabcolsep}{.35em}
\resizebox{\linewidth}{!}{
\begin{tabular}{clc cc c cc c cc c c c c c c}
\toprule
& & & \multicolumn{2}{c}{\textbf{Flickr8k-Expert}} & & \multicolumn{2}{c}{\textbf{Flickr8k-CF}} & & \multicolumn{2}{c}{\textbf{VATEX-EVAL}} & & \textbf{Pascal-50S} & & \textbf{FOIL} & & \textbf{ActivityNet-FOIL} \\
\cmidrule{4-5} \cmidrule{7-8} \cmidrule{10-11} \cmidrule{13-13} \cmidrule{15-15} \cmidrule{17-17}
& & & Kendall $\tau_b$ & Kendall $\tau_c$  & & Kendall $\tau_b$ & Kendall $\tau_c$ & & Kendall $\tau_b$ & Spearman $\rho$ & & Accuracy & & Accuracy & & Accuracy \\
\midrule
& CLIP-S~\cite{hessel2021clipscore} &  &  51.7 &  52.1 &  &  34.9 &  18.0 &  &  - &  - &  &  81.1 &  &  90.6 &  &  - \\
& EMScore~\cite{shi2022emscore}  & & - & - & & - & - & & 24 1& 31.4 & & - & & - & & 90.0 \\
\rowcolor{LightCyan}
\cellcolor[rgb]{1,1,1} & &  &  \textbf{54.5} &  \textbf{54.9} &  &  \textbf{35.9} &  \textbf{18.5} &  & \textbf{26.8} & \textbf{34.7} &  &  \textbf{82.9} &  &  \textbf{91.1} &  &  \textbf{90.7} \\
\rowcolor{LightCyan}
\cellcolor[rgb]{1,1,1} \multirow{-4}{*}{\textbf{CLIP ViT-B/16}} & \multirow{-2}{*}{\textbf{\ours}} & & (\textcolor{blue}{+2.8}) & (\textcolor{blue}{+2.8}) & & (\textcolor{blue}{+1.0}) & (\textcolor{blue}{+0.5}) & & (\textcolor{blue}{+2.7}) & (\textcolor{blue}{+3.3}) & & (\textcolor{blue}{+1.8}) & & (\textcolor{blue}{+0.5}) & & (\textcolor{blue}{+0.7}) \\
\midrule
 & CLIP-S~\cite{hessel2021clipscore} & & 52.6 & 53.0 & & 35.2 & 18.2 & & - & - & & 81.7 & & 90.9 & & - \\
& EMScore~\cite{shi2022emscore}  & & - & - & & - & - & & 26.7 & 34.7 & & - & & - & & 89.0 \\
\rowcolor{LightCyan}
\cellcolor[rgb]{1,1,1} & &  &  \textbf{55.1} &  \textbf{55.5} &  &  \textbf{36.8} &  \textbf{19.0} &  &  \textbf{28.9} &  \textbf{37.4} &  &  \textbf{82.2} &  &  \textbf{91.9} &  &  \textbf{91.2} \\
\rowcolor{LightCyan}
\cellcolor[rgb]{1,1,1} \multirow{-4}{*}{\textbf{CLIP ViT-L/14}} & \multirow{-2}{*}{\textbf{\ours}} & & (\textcolor{blue}{+2.5}) & (\textcolor{blue}{+2.5}) & & (\textcolor{blue}{+1.6}) & (\textcolor{blue}{+0.8}) & & (\textcolor{blue}{+2.2}) & (\textcolor{blue}{+2.7}) & & (\textcolor{blue}{+0.5}) & & (\textcolor{blue}{+1.0}) & & (\textcolor{blue}{+2.2}) \\
\midrule
& CLIP-S~\cite{hessel2021clipscore} & & 52.3 & 52.6 & & 35.4 & 18.3 & & - & - & & 81.2 & & 88.9 & & - \\
\textbf{OpenCLIP} & EMScore~\cite{shi2022emscore} & & - & - & & - & - & & 24.8 & 32.2 & & - & & - & & 88.2 \\
\rowcolor{LightCyan}
\cellcolor[rgb]{1,1,1} \textbf{ViT-B/32} & &  &  \textbf{53.6} &  \textbf{53.9} &  &  \textbf{36.1} &  \textbf{18.6} &  & \textbf{25.4} & \textbf{33.1} &  &  \textbf{82.1} &  &  \textbf{90.1} &  &  \textbf{89.5} \\
\rowcolor{LightCyan}
\cellcolor[rgb]{1,1,1} & \multirow{-2}{*}{\textbf{\ours}} & & (\textcolor{blue}{+1.3}) & (\textcolor{blue}{+1.3}) & & (\textcolor{blue}{+0.7}) & (\textcolor{blue}{+0.3}) & & (\textcolor{blue}{+0.6}) & (\textcolor{blue}{+0.9}) & & (\textcolor{blue}{+0.9}) & & (\textcolor{blue}{+1.2}) & & (\textcolor{blue}{+1.3}) \\
\midrule
& CLIP-S~\cite{hessel2021clipscore}  & & 54.4 & 54.5 & & 36.6 & 18.9 & & - & - & & 82.5 & & 92.2 & & - \\
\textbf{OpenCLIP} & EMScore~\cite{shi2022emscore}  & & - & - & & - & - & & 27.0 & 35.0 & & - & & - & & 90.7 \\
\rowcolor{LightCyan}
\cellcolor[rgb]{1,1,1} \textbf{ViT-L/14} &  &  &  \textbf{55.3} &  \textbf{55.7} &  &  \textbf{37.0} &  \textbf{19.1} &  &  \textbf{27.8} & \textbf{36.1} &  &  \textbf{82.8} &  &  \textbf{93.1} &  &  \textbf{91.2} \\
\rowcolor{LightCyan}
\cellcolor[rgb]{1,1,1} & \multirow{-2}{*}{\textbf{\ours}} & & (\textcolor{blue}{+0.9}) & (\textcolor{blue}{+1.2}) & & (\textcolor{blue}{+0.4}) & (\textcolor{blue}{+0.2}) & & (\textcolor{blue}{+0.8}) & (\textcolor{blue}{+1.1}) & & (\textcolor{blue}{+0.3}) & & (\textcolor{blue}{+0.9}) & & (\textcolor{blue}{+0.5}) \\
\bottomrule
\end{tabular}
}
 \vspace{-0.1cm}
\caption{Human correlation and accuracy scores on both image and video captioning datasets using different cross-modal backbones.}
\label{tab:features}
\vspace{-0.3cm}
\end{table*}

\subsection{Sensitivity to object hallucination}
Correctly identifying captions with potential object hallucinations (\ie~with objects that are not present in the image or video) is fundamental for the captioning task~\cite{rohrbach2018object}. Therefore, we extend our analysis to two datasets for detecting hallucinations in textual sentences, namely FOIL~\cite{shekhar2017foil} and ActivityNet-FOIL~\cite{shi2022emscore}. In particular, the FOIL dataset is composed of image-caption pairs from the COCO dataset~\cite{lin2014microsoft}. In this case, captions are perturbed by creating modified versions that are highly similar to the original ones but contain one single error (\ie~a foil word). For a fair comparison, we take the subset of the validation set that does not overlap with the portion of COCO used to finetune our model thus obtaining 8k images, each associated with a foil-correct textual pair. The ActivityNet-FOIL dataset, instead, contains video-text pairs from the ActivityNet test set~\cite{zhou2019grounded}. Each video comes with two annotated paragraphs, one used to construct foil-correct pair and the other used as ground-truth for reference-based metrics. To create a foil caption, a noun phrase in the original caption is replaced with a similar but incorrect visual concept. Overall, the dataset is composed of 1,900 foil-correct paragraph pairs.  

Since each image or video is associated with a foil-correct caption pair, we measure the portion of times in which the correct caption obtains a higher score than the foil one. Table~\ref{tab:foil} shows the accuracy results on the considered datasets. As it can be seen, \ours achieves better results than previous solutions, increasing the accuracy score of 2.7 and 0.6 points compared to CLIP-S and EMScore, respectively. Similar improvements can also be observed in the reference-based version, demonstrating the capabilities of our metric to correctly identify hallucinated objects.

\subsection{System-level correlation}
After demonstrating the benefits of using \ours over other evaluation metrics, we also analyze its effectiveness when evaluating existing captioning methods. To this aim, we consider different popular captioning models and compute their predictions on images coming from the COCO test set. Results are reported in Table~\ref{tab:sota} in terms of BLEU-4, METEOR, CIDEr, CLIP-S, and our \ours, in both reference-free and reference-based versions. We also include the results of a human-based baseline, in which for each sample one human-annotated sentence (among the five provided by the COCO dataset) is randomly selected as candidate caption and compared with the remaining references\footnote{The BLEU-4 score of the human-based baseline is not reported due to its sensitivity to the number of references used for evaluation.}. As shown in the table, our metric well correlates with previous ones in identifying the best captioning model. Interestingly, \ours can also effectively evaluate human-annotated sentences, unlike for example the METEOR and CIDEr scores which rank human captions even below those generated by early captioning approaches~\cite{xu2015show,vinyals2015show}.

\subsection{Analyzing other cross-modal features}\label{sec:diffFeats}
Finally, we report in Table~\ref{tab:features} captioning evaluation results when using different cross-modal features. In particular, we employ ViT-B/16 and ViT-L/14 models of CLIP~\cite{radford2021learning} and the ViT-B/32 and ViT-L/14 versions of the open source implementation (\ie~OpenCLIP~\cite{wortsman2022robust}\footnote{\url{https://github.com/mlfoundations/open_clip}}) trained on the English subset of the LAION-5B dataset~\cite{schuhmann2022laion}. For all backbones, we employ the same finetuning procedure and training settings described in Sec~\ref{sec:details}. We conduct the analysis on the majority of the datasets considered in the previous experiments and compare the proposed \ours with CLIP-S and EMScore, respectively for image and video captioning datasets. Noticeably, \ours achieves the best results across all cross-modal backbones and all datasets, overcoming correlation and accuracy scores of other metrics by a large margin. When comparing the results when using different backbones, both ViT-L/14 models outperform other considered architectures as well as the standard CLIP ViT-B/32 model used in previous experiments, thus demonstrating the usefulness of using more powerful cross-modal models to evaluate captioning predictions.

%% file: sections/05_conclusion.tex
In this paper, we have proposed a positive-augmented contrastive learning approach for image and video captioning evaluation. Our proposal, PAC-S, is trained by considering cleaned data sources and leveraging synthetic images and captions as an additional source of supervision. Experimentally, we have demonstrated that PAC-S is superior to all previous metrics in terms of correlation with human judgment and sensitivity to hallucinated objects in both reference-free and reference-based settings.

%% file: supplementary.tex
\appendix

\section{Additional Experimental Results}
\tinytit{Correlation with MID score}
In addition to the experiments presented in the main paper, we conducted further comparisons with the MID metric~\cite{kim2022mutual}. Since it exploits CLIP-based features as CLIP-S~\cite{hessel2021clipscore} and our proposal, in Table~\ref{tab:mid} we compare the results of the original MID score with a re-implemented version that uses our embeddings in place of those of CLIP. In particular, we conduct this analysis on the Flickr8k-Expert, Flickr8k-CF, and FOIL datasets and show that using our embeddings can further improve the results of the MID score in the majority of the considered settings, thus further demonstrating the appropriateness of our positive-augmented contrastive learning approach.

\tit{Reference-based results using ViT-based backbones} As a complement to Table~\ref{tab:features}, in Table~\ref{tab:featuresViT} we report the referenced-based results using different cross-modal features. In particular, we experiment with different ViT-based backbones of CLIP~\cite{radford2021learning} and OpenCLIP~\cite{wortsman2022robust} models. From these results, we confirm the effectiveness of \ours also in the reference-based setting on both image and video captioning datasets. Both ViT-L/14 models outperform the others even in this case, still confirming that using more powerful features can lead to better results.

\begin{table}[b]
\centering
\setlength{\tabcolsep}{.2em}
\resizebox{\linewidth}{!}{
\begin{tabular}{lcc cc c cc c c c c}
\toprule
 & & & \multicolumn{2}{c}{\textbf{Flickr8k-Expert}} & & \multicolumn{2}{c}{\textbf{Flickr8k-CF}}  & & \textbf{Pascal-50S} & & \textbf{FOIL} \\
  \cmidrule{4-5} \cmidrule{7-8} \cmidrule{10-10} \cmidrule{12-12}
 & \textbf{Features} & & Kendall $\tau_b$ & Kendall $\tau_c$ & & Kendall $\tau_b$ & Kendall $\tau_c$ & & Accuracy & & Accuracy \\
\midrule
MID~\cite{kim2022mutual} & CLIP & & - & 54.9 & & \textbf{37.3} & - & & \textbf{85.2} & & 90.5 \\
\midrule
MID$^\dagger$ & CLIP & & 54.3 & 54.6 & & 36.5 & 18.7 & & 84.6 & & 93.2 \\
\rowcolor{LightCyan}
\textbf{MID$^\dagger$} & \textbf{Ours} & &  {\textbf{54.7}} & \textbf{55.1} & & 36.7 & \textbf{18.8} & & 85.0 & & \textbf{93.3} \\
\bottomrule
\end{tabular}
}
\caption{Performance of MID with CLIP and PAC ViT-B/32 features. The $\dagger$ marker indicates our re-implementation.}
\label{tab:mid}
\vspace{-.15cm}
\end{table}

\begin{figure}[b]
\centering
\setlength{\tabcolsep}{.1em}
\begin{tabular}{ccc}
\includegraphics[width=0.325\linewidth]{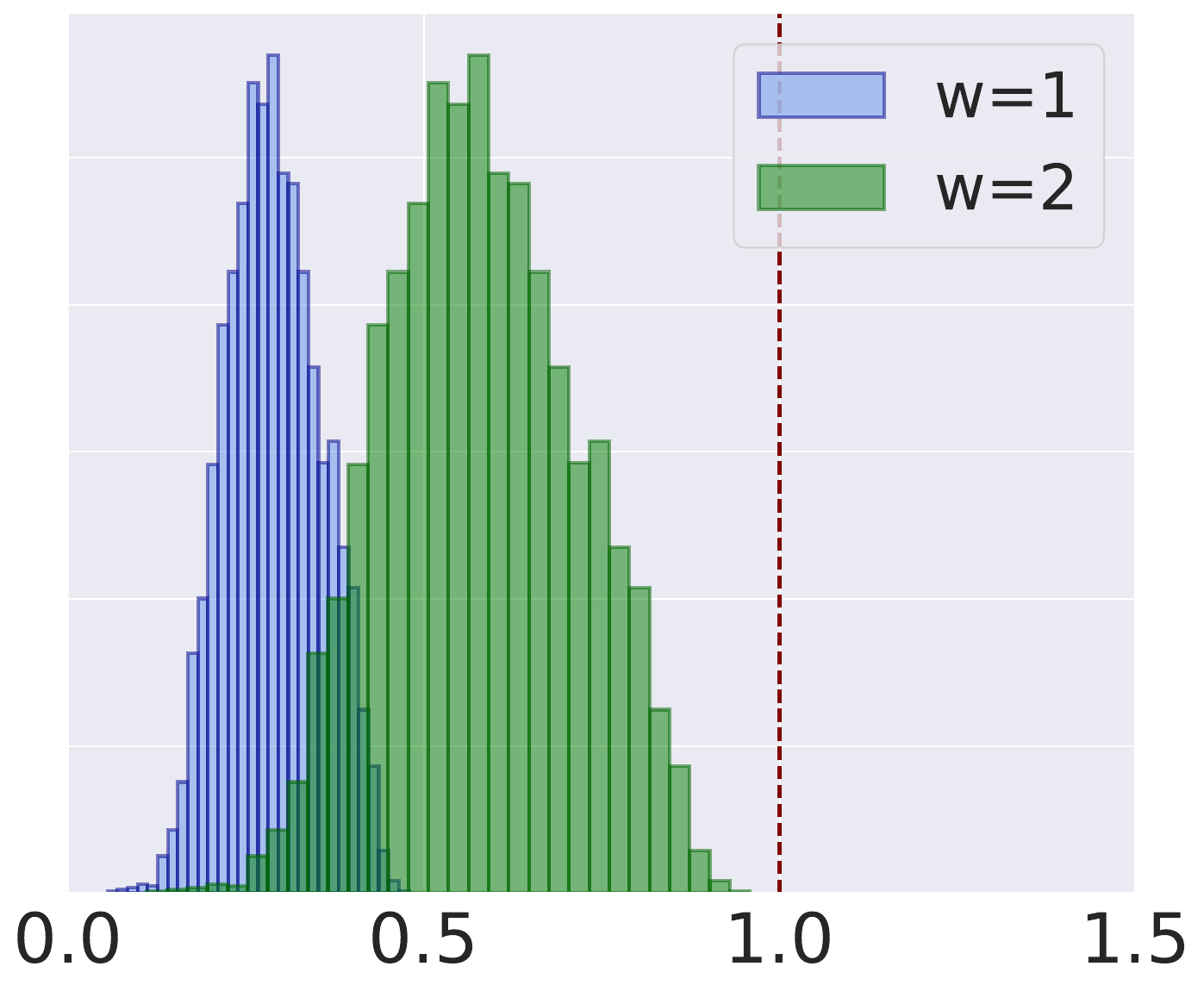} &
\includegraphics[width=0.325\linewidth]{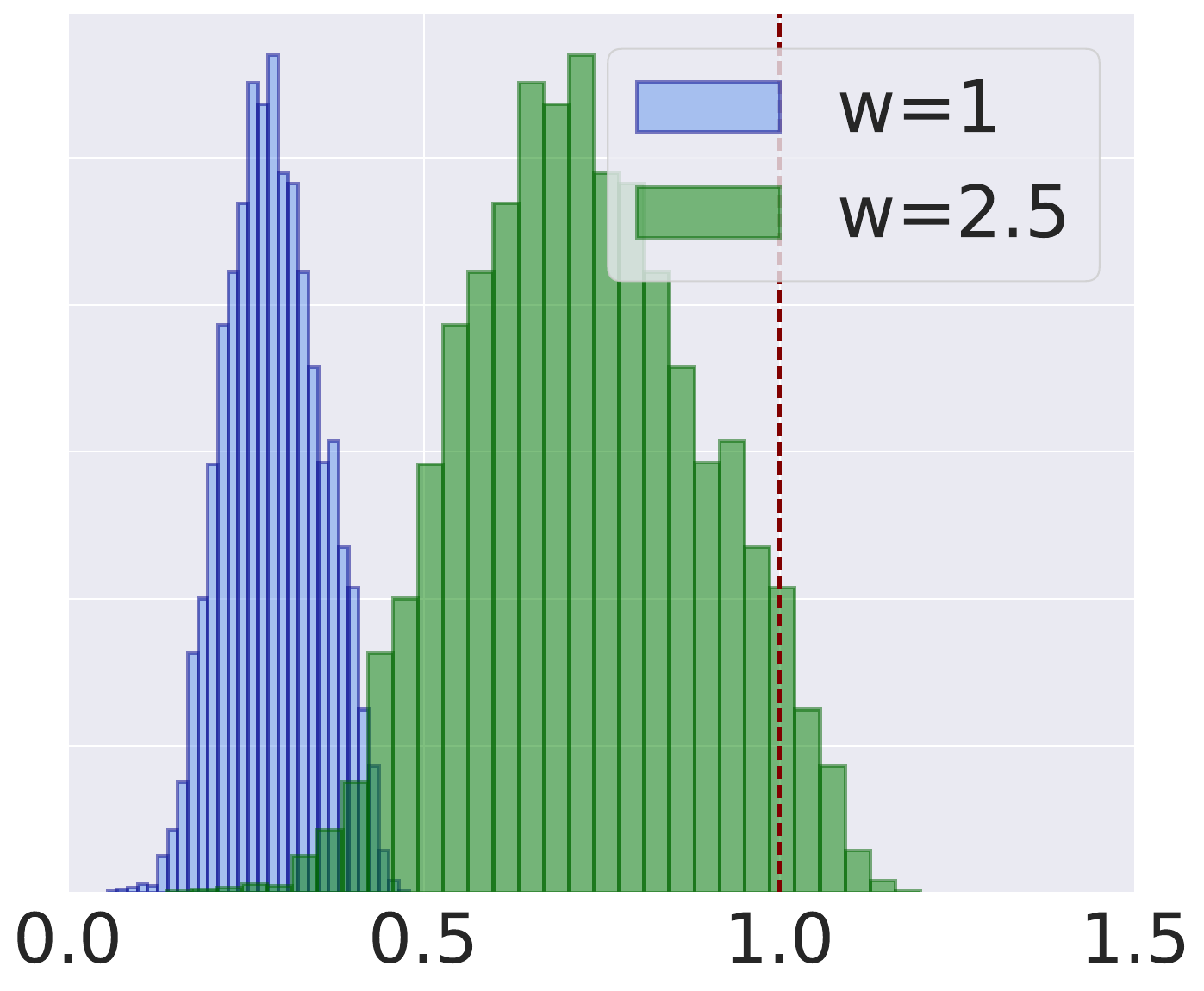} &
\includegraphics[width=0.325\linewidth]{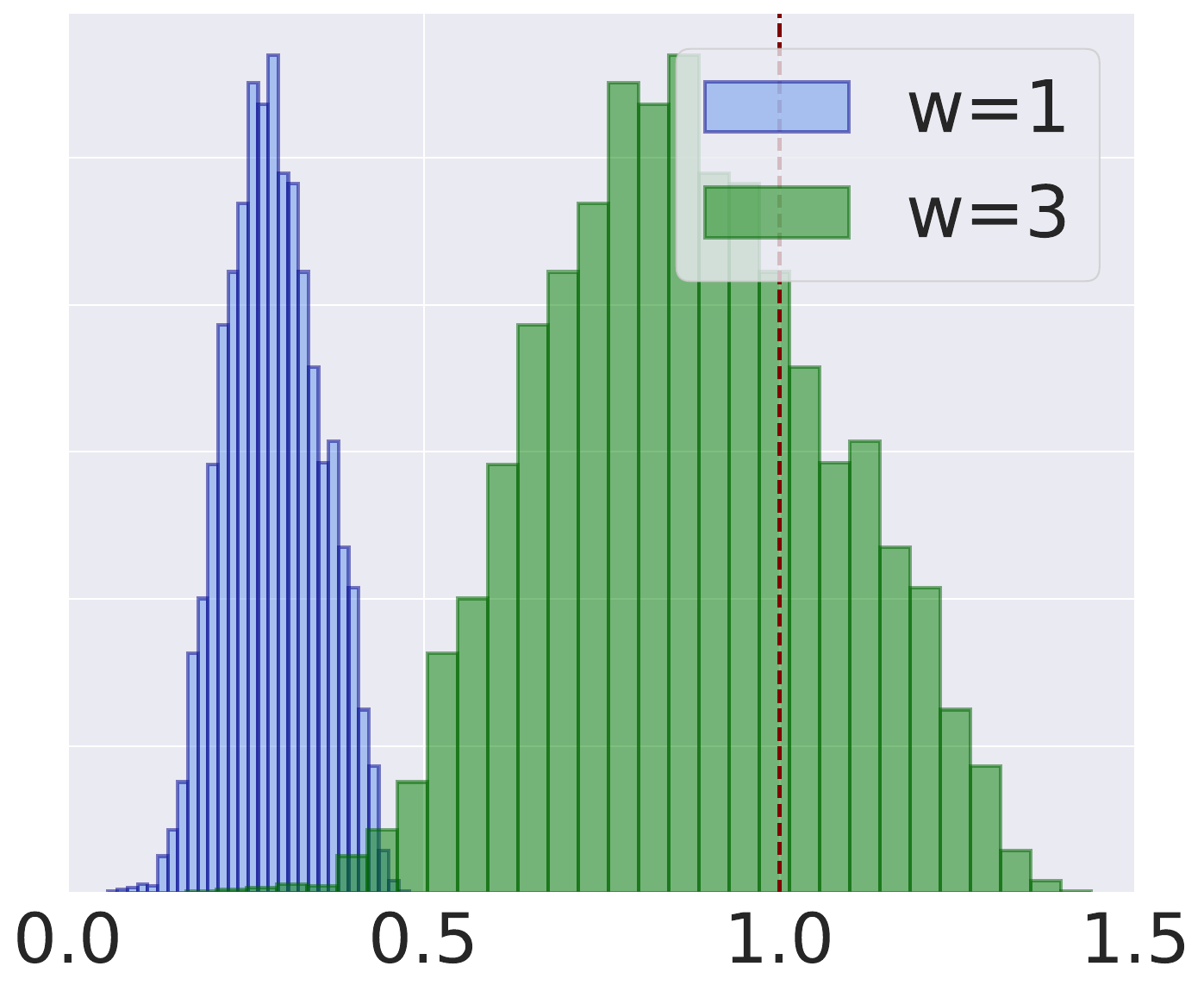}
\end{tabular}
\vspace{-.3cm}
\caption{Distribution of PAC scores using different $w$ (Eq.~\ref{eq:clip_score} of the main paper).}
\label{fig:graph}
\vspace{-.3cm}
\end{figure}

\tit{Analyzing ResNet-based backbones}
In Table~\ref{tab:featuresResnet}, we conduct the same analysis in both reference-free and reference-based settings but using visual features extracted from a ResNet backbone~\cite{he2016deep}. Specifically, we use the following CLIP-based models: ResNet-50, ResNet-101, and ResNet-50$\times$4, which employ an EfficientNet-style architecture scaling. For these experiments, we finetune the last attention pooling of the visual backbone and the final projection of the textual branch using the same settings described in the main paper. Also in this case, our metric achieves the best results in almost all datasets, with the only exception of VATEX-EVAL in which the EMScore obtains slightly better correlation scores. 

\begin{table*}[t]
\small
\centering
\setlength{\tabcolsep}{.35em}
\resizebox{\linewidth}{!}{
\begin{tabular}{clc cc c cc c cc c c c c c c}
\toprule
& & & \multicolumn{2}{c}{\textbf{Flickr8k-Expert}} & & \multicolumn{2}{c}{\textbf{Flickr8k-CF}} & & \multicolumn{2}{c}{\textbf{VATEX-EVAL}} & & \textbf{PASCAL-50S} & & \textbf{FOIL} & & \textbf{ActivityNet-FOIL} \\
\cmidrule{4-5} \cmidrule{7-8} \cmidrule{10-11} \cmidrule{13-13} \cmidrule{15-15} \cmidrule{17-17}
& & & Kendall $\tau_b$ & Kendall $\tau_c$  & & Kendall $\tau_b$ & Kendall $\tau_c$ & & Kendall $\tau_b$ & Spearman $\rho$ & & Accuracy & & Accuracy & & Accuracy \\
\midrule
& RefCLIP-S~\cite{hessel2021clipscore} & & 53.6 & 54.0 & & 36.7 & 19.0 & & - & - & & 84.0 & & 94.8 & & - \\
& EMScoreRef~\cite{shi2022emscore}  & & - & - & & - & - & & 37.1 & 47.5 & & - & & - & & 92.2 \\
\rowcolor{LightCyan}
\cellcolor[rgb]{1,1,1} & &  &  \textbf{56.0} &  \textbf{56.4} &  &  \textbf{37.5} &  \textbf{19.4} &  & \textbf{38.8} & \textbf{49.6} &  &  \textbf{84.8} &  &  \textbf{95.1} &  &  \textbf{92.6} \\
\rowcolor{LightCyan}
\cellcolor[rgb]{1,1,1} \multirow{-4}{*}{\textbf{CLIP ViT-B/16}} & \multirow{-2}{*}{\textbf{\oursref}} & & (\textcolor{blue}{+2.4}) & (\textcolor{blue}{+2.4}) & & (\textcolor{blue}{+0.8}) & (\textcolor{blue}{+0.4}) & & (\textcolor{blue}{+1.7}) & (\textcolor{blue}{+2.1}) & & (\textcolor{blue}{+0.8}) & & (\textcolor{blue}{+0.3}) & & (\textcolor{blue}{+0.4}) \\
\midrule
 & RefCLIP-S~\cite{hessel2021clipscore} & & 54.0 & 54.4 & & 36.5 & 18.9 & & - & - & & \textbf{85.0} & & 94.9 & & - \\
& EMScoreRef~\cite{shi2022emscore}  & & - & - & & - & - & & 37.0 & 47.4 & & - & & - & & 93.5 \\
\rowcolor{LightCyan}
\cellcolor[rgb]{1,1,1} & & &  \textbf{56.7} &  \textbf{57.1} &  &  \textbf{37.7} &  \textbf{19.5} &  &  \textbf{38.6} &  \textbf{49.3} &  &  \textbf{85.0} &  &  \textbf{95.3} &  &  \textbf{94.2} \\
\rowcolor{LightCyan}
\cellcolor[rgb]{1,1,1} \multirow{-4}{*}{\textbf{CLIP ViT-L/14}} & \multirow{-2}{*}{\textbf{\oursref}} & & (\textcolor{blue}{+2.7}) & (\textcolor{blue}{+2.7}) & & (\textcolor{blue}{+1.2}) & (\textcolor{blue}{+0.6}) & & (\textcolor{blue}{+1.6}) & (\textcolor{blue}{+1.9}) & & (+0.0) & & (\textcolor{blue}{+0.4}) & & (\textcolor{blue}{+0.7}) \\
\midrule
 & RefCLIP-S~\cite{hessel2021clipscore} & & 53.9 & 54.3 & & 36.8 & 19.0 & & - & - & & \textbf{84.7} & & \textbf{94.7} & & - \\
\textbf{OpenCLIP} & EMScoreRef~\cite{shi2022emscore} & & - & - & & - & - & & 38.4 & 49.1 & & - & & - & & 93.0 \\
\rowcolor{LightCyan}
\cellcolor[rgb]{1,1,1} \textbf{ViT-B/32} &  &  &  \textbf{54.8} &  \textbf{55.2} &  &  \textbf{37.4} &  \textbf{19.3} &  &   \textbf{38.8} &   \textbf{49.5} &  & 84.5 &  & 94.1 &  &  \textbf{93.6} \\
\rowcolor{LightCyan}
\cellcolor[rgb]{1,1,1} & \multirow{-2}{*}{\textbf{\oursref}} & & (\textcolor{blue}{+0.9}) & (\textcolor{blue}{+0.9}) & & (\textcolor{blue}{+0.6}) & (\textcolor{blue}{+0.3}) & & (\textcolor{blue}{+0.4}) & (\textcolor{blue}{+0.4}) & & (\textcolor{black}{-0.2}) & & (\textcolor{black}{-0.6}) & & (\textcolor{blue}{+0.6}) \\
\midrule
 & RefCLIP-S~\cite{hessel2021clipscore}  & & 55.7 & 55.8 & & 37.5 & 19.4 & & - & - & & \textbf{85.3} & & \textbf{95.9} & & - \\
\textbf{OpenCLIP} & EMScoreRef~\cite{shi2022emscore}  & & - & - & & - & - & & 39.4 & 50.3 & & - & & - & & 94.0 \\
\rowcolor{LightCyan}
\cellcolor[rgb]{1,1,1} \textbf{ViT-L/14}  &  &  &  \textbf{56.5} &  \textbf{56.9} &  &  \textbf{38.0} &  \textbf{19.7} &  &  \textbf{40.3} &  \textbf{51.4} &  &  84.9 &  &  95.8 &  &  \textbf{94.4} \\
\rowcolor{LightCyan}
\cellcolor[rgb]{1,1,1} & \multirow{-2}{*}{\textbf{\oursref}} & & (\textcolor{blue}{+0.8}) & (\textcolor{blue}{+1.1}) & & (\textcolor{blue}{+0.5}) & (\textcolor{blue}{+0.3}) & & (\textcolor{blue}{+0.9}) & (\textcolor{blue}{+1.1}) & & (-0.4) & & (\textcolor{black}{-0.1}) & & (\textcolor{blue}{+0.4}) \\
\bottomrule
\end{tabular}
}
\vspace{-.1cm}
\caption{Captioning evaluation results in a reference-based setting on both image and video captioning datasets using different cross-modal features.}
\label{tab:featuresViT}
\vspace{-.15cm}
\end{table*}

\begin{table*}[t]
\small
\centering
\setlength{\tabcolsep}{.35em}
\resizebox{\linewidth}{!}{
\begin{tabular}{clc cc c cc c cc c c c c c c}
\toprule
& & & \multicolumn{2}{c}{\textbf{Flickr8k-Expert}} & & \multicolumn{2}{c}{\textbf{Flickr8k-CF}} & & \multicolumn{2}{c}{\textbf{VATEX-EVAL}} & & \textbf{PASCAL-50S} & & \textbf{FOIL} & & \textbf{ActivityNet-FOIL} \\
\cmidrule{4-5} \cmidrule{7-8} \cmidrule{10-11} \cmidrule{13-13} \cmidrule{15-15} \cmidrule{17-17}
& & & Kendall $\tau_b$ & Kendall $\tau_c$  & & Kendall $\tau_b$ & Kendall $\tau_c$ & & Kendall $\tau_b$ & Spearman $\rho$ & & Accuracy & & Accuracy & & Accuracy \\
\midrule
 & CLIP-S~\cite{hessel2021clipscore} & & 51.0 & 51.4 & & 34.0 & 17.6 & & - & - & & 80.6 & & \textbf{87.9} & & - \\
& EMScore~\cite{shi2022emscore} & & - & - & & - & - & & \textbf{22.0} & \textbf{28.6} & & - & & - & & 87.0 \\
\rowcolor{LightCyan}
\cellcolor[rgb]{1,1,1} & & & \textbf{52.6} & \textbf{52.9} & &  \textbf{34.6} & \textbf{17.9} &  & 19.4 & 25.4 &  & \textbf{81.7} &  & 87.1 &  &  \textbf{87.7} \\
\rowcolor{LightCyan}
\cellcolor[rgb]{1,1,1} \multirow{-4}{*}{\textbf{CLIP RN50}} & \multirow{-2}{*}{\textbf{\ours}} & & (\textcolor{blue}{+1.6}) & (\textcolor{blue}{+1.5}) & & (\textcolor{blue}{+0.6}) & (\textcolor{blue}{+0.3}) & & (-2.6) & (-3.2) & & (\textcolor{blue}{+1.1}) & & (-0.8) & & (\textcolor{blue}{+0.7}) \\
\midrule & RefCLIP-S~\cite{hessel2021clipscore} & & 52.5 & 52.8 & & 35.9 & 18.5 & & - & - & & 83.4 & & \textbf{93.4} & & - \\
& EMScoreRef~\cite{shi2022emscore} & & - & - & & - & - & & \textbf{36.6 } & {\textbf{46.9}}  & & - & & - & & 91.8 \\
\rowcolor{LightCyan}
\cellcolor[rgb]{1,1,1} & & &  \textbf{54.1} & \textbf{54.5} &  & \textbf{36.4}  & {\textbf{18.8}}  &  &  36.4 &  46.7 &  & \textbf{83.8} &  & 93.1  &  &  {\textbf{92.7}} \\
\rowcolor{LightCyan}
\cellcolor[rgb]{1,1,1} \multirow{-4}{*}{\textbf{CLIP RN50}} & \multirow{-2}{*}{\textbf{\oursref }} & & (\textcolor{blue}{+1.6}) & (\textcolor{blue}{+1.7}) & & (\textcolor{blue}{+0.5}) & (\textcolor{blue}{+0.3}) & & (-0.2) & (-0.2) & & (\textcolor{blue}{+0.4}) & & (\textcolor{black}{-0.3}) & & (\textcolor{blue}{+0.9}) \\
\midrule
 & CLIP-S~\cite{hessel2021clipscore} & & 50.5 & 50.9 & & 33.5 & 17.3 & & - & - & & 80.5 & & \textbf{89.1} & & - \\
& EMScore~\cite{shi2022emscore} & & - & - & & - & - & & \textbf{21.6} & \textbf{28.2} & & - & & - & & \textbf{89.6} \\
\rowcolor{LightCyan}
\cellcolor[rgb]{1,1,1} & & & \textbf{53.4} &  \textbf{53.7} &  &  \textbf{34.4} &  \textbf{17.8} &  &  20.4 &  26.6 &  &  \textbf{81.8} &  &  89.0 &  &  88.9 \\
\rowcolor{LightCyan}
\cellcolor[rgb]{1,1,1} \multirow{-4}{*}{\textbf{CLIP RN101}} & \multirow{-2}{*}{\textbf{\ours}} & & (\textcolor{blue}{+2.9}) & (\textcolor{blue}{+2.8}) & & (\textcolor{blue}{+0.9}) & (\textcolor{blue}{+0.5}) & & (-1.2) & (-1.6) & & (\textcolor{blue}{+1.3}) & & (-0.1) & & (-0.7) \\
\midrule
& RefCLIP-S~\cite{hessel2021clipscore} & & 52.2 & 52.6 & & 35.6 & 18.4 & & - & - & & 83.3 & & 95.2 & & - \\
& EMScoreRef~\cite{shi2022emscore}& & - & - & & - & - & & 36.6 & 46.9  & & - & & - & & 91.7 \\
\rowcolor{LightCyan}
\cellcolor[rgb]{1,1,1} & & & {\textbf{55.5}}  &   {\textbf{55.9}}  &  &  {\textbf{36.6}}  &  {\textbf{ 18.9}} &  &  {\textbf{37.1}}  &  {\textbf{47.5}}  &  &  {\textbf{84.8}}  &  &  {\textbf{95.4}}  &  & {\textbf{92.1}}  \\
\rowcolor{LightCyan}
\cellcolor[rgb]{1,1,1} \multirow{-4}{*}{\textbf{CLIP RN101}} & \multirow{-2}{*}{\textbf{\oursref}} & & (\textcolor{blue}{+3.3}) & (\textcolor{blue}{+3.3}) & & (\textcolor{blue}{+1.0}) & (\textcolor{blue}{+0.5}) & & (\textcolor{blue}{+0.5}) & (\textcolor{blue}{+0.6}) & & (\textcolor{blue}{+1.5}) & & (\textcolor{blue}{+0.2}) & & (\textcolor{blue}{+0.4}) \\
\midrule
& CLIP-S~\cite{hessel2021clipscore} & & 50.7 & 51.0 & & 34.0 & 17.6 & & - & - & & 80.7 & & 89.5 & & - \\
& EMScore~\cite{shi2022emscore}& & - & - & & - & - & & \textbf{22.0} & \textbf{28.8}  & & - & & - & & \textbf{88.8} \\
\rowcolor{LightCyan}
\cellcolor[rgb]{1,1,1} & & & \textbf{53.9} &  \textbf{54.3} &  &  \textbf{35.9} &  \textbf{18.6} &  & 21.9 & 28.6 &  &  \textbf{82.5} &  &  \textbf{90.5} &  &  87.7 \\
\rowcolor{LightCyan}
\cellcolor[rgb]{1,1,1} \multirow{-4}{*}{\textbf{CLIP RN50$\times$4}} & \multirow{-2}{*}{\textbf{\ours}} & & (\textcolor{blue}{+3.2}) & (\textcolor{blue}{+3.3}) & & (\textcolor{blue}{+1.9}) & (\textcolor{blue}{+1.0}) & & (-0.1) & (-0.2) & & (\textcolor{blue}{+1.8}) & & (\textcolor{blue}{+1.0}) & & (-1.1) \\
\midrule
& RefCLIP-S~\cite{hessel2021clipscore} & & 52.3 & 52.7 & & 36.1 & 18.7 & & - & - & & 83.3 & & 95.3 & & - \\
& EMScoreRef~\cite{shi2022emscore} & & - & - & & - & - & & 36.7 & 45.0 & & - & & - & & 91.5 \\
\rowcolor{LightCyan}
\cellcolor[rgb]{1,1,1} & &  & {\textbf{56.2}}  &  {\textbf{56.6}} &  & {\textbf{37.3}}  & {\textbf{19.3}}  &  & {\textbf{37.4}}  & {\textbf{47.7}}  &  & {\textbf{84.8}}  &  &  {\textbf{95.8}}  &  &  \textbf{91.9} \\
\rowcolor{LightCyan}
\cellcolor[rgb]{1,1,1} \multirow{-4}{*}{\textbf{CLIP RN50$\times$4}} & \multirow{-2}{*}{\textbf{\oursref}} & & (\textcolor{blue}{+3.9}) & (\textcolor{blue}{3.9}) & & (\textcolor{blue}{+1.2}) & (\textcolor{blue}{+0.6}) & & (\textcolor{blue}{+0.7}) & (\textcolor{blue}{+2.7}) & & (\textcolor{blue}{+1.5}) & & (\textcolor{blue}{+0.5}) & & (\textcolor{blue}{+0.4}) \\
\midrule
\end{tabular}
}
\vspace{-.1cm}
\caption{Additional human correlation and accuracy scores on both image and video captioning datasets using different cross-modal ResNet-based backbones.}
\label{tab:featuresResnet}
\vspace{-.35cm}
\end{table*}

\tit{Choice of hyperparameters}
The scaling factor, denoted by $w$ in Eq.~\ref{tab:featuresViT}, is utilized to adjust the scale of the final metric to improve its numerical readability, without affecting the ranking of the results. CLIP-S also employs a comparable technique, where $w$ is assigned the value of 2.5.
To provide additional clarification, we present in Fig.~\ref{fig:graph} the impact of varying values of $w$. The raw PAC-S scores with $w=1$ lie between 0 and 0.5 on all datasets. Therefore, we decide to use a scaling factor $w$ equal to 2 which stretch the PAC-S scores between 0 and 1. 

\begin{figure*}[t]
\centering
\includegraphics[width=\linewidth]{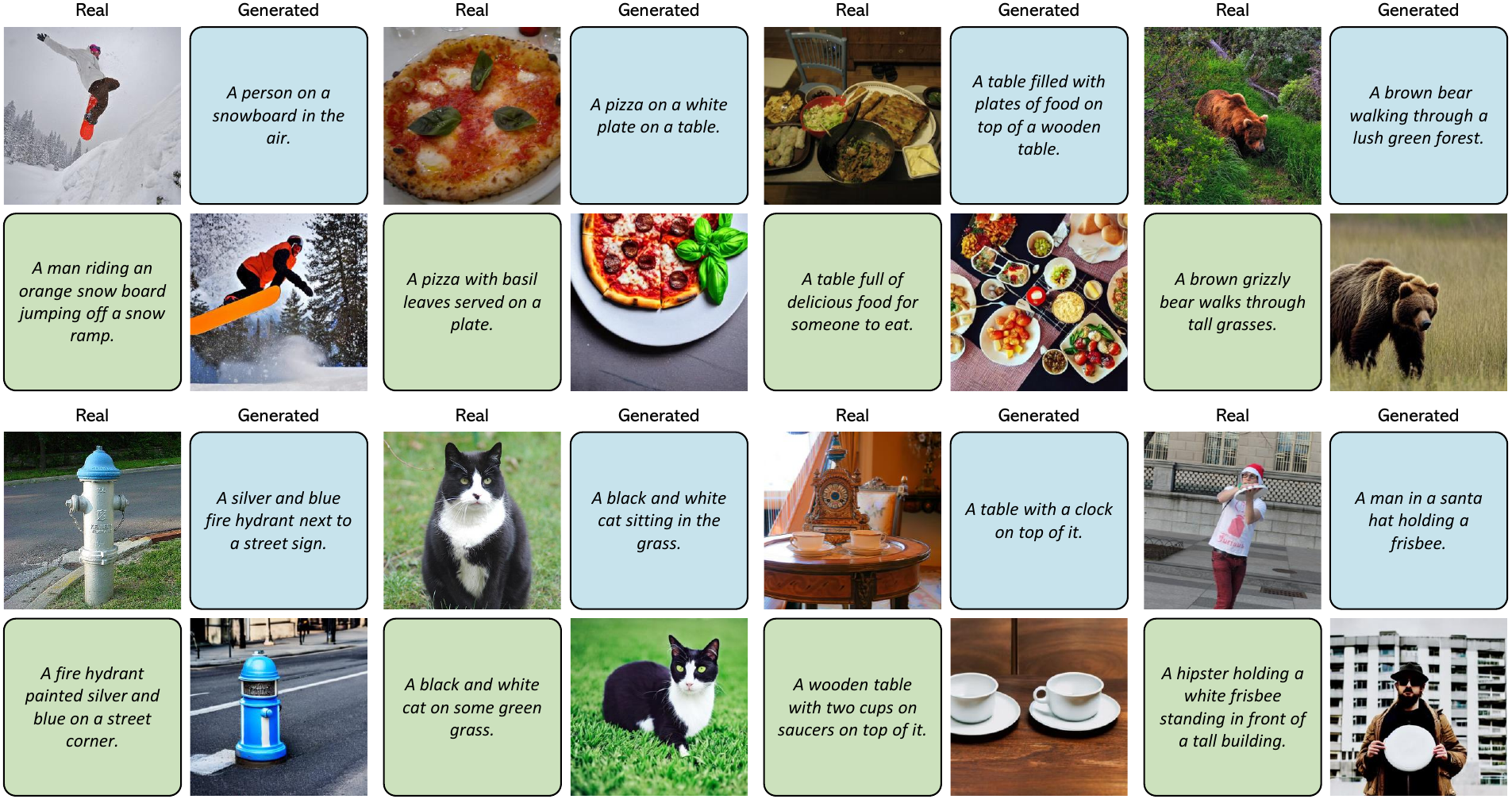}
\caption{Additional real and generated image-text samples used to augment the training set for positive-augmented contrastive learning.}
\label{fig:generated}
\vspace{-.15cm}
\end{figure*}

\begin{figure*}[t]
\centering
\includegraphics[width=\linewidth]{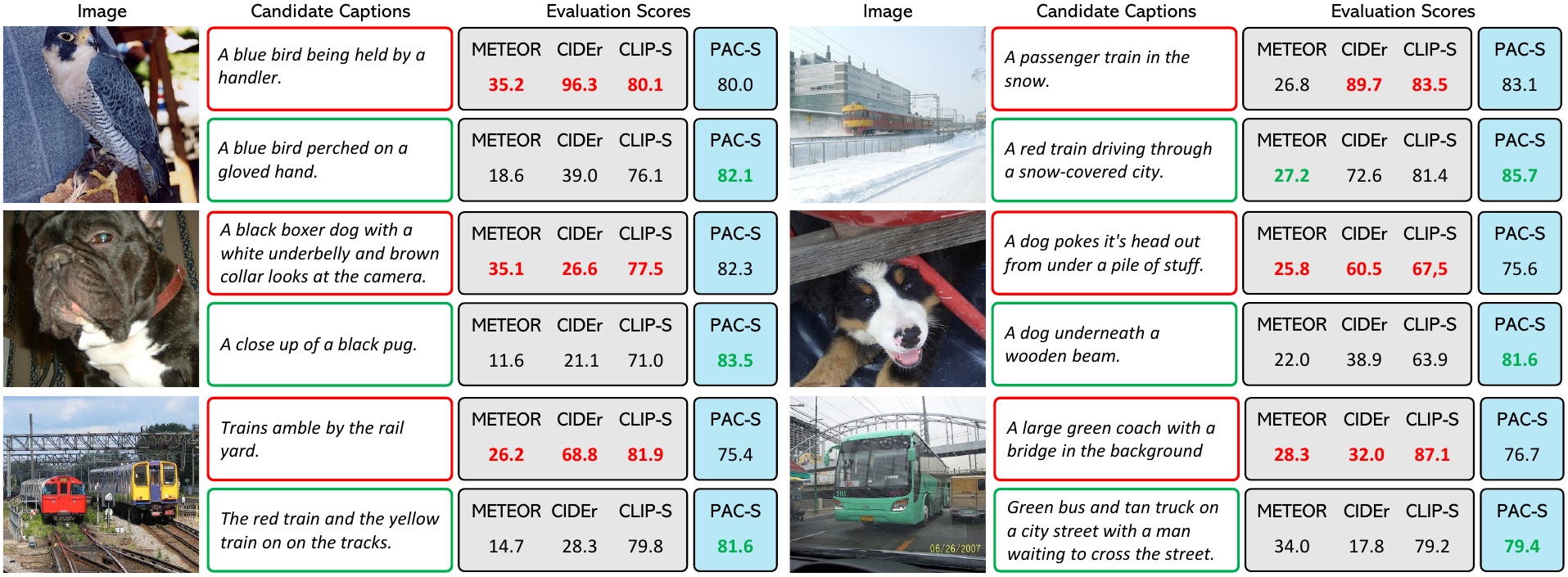}
\caption{Additional comparisons of existing metrics for captioning with respect to \ours on the Pascal-50S dataset. The candidate caption highlighted in green is the one preferred by humans.}
\label{fig:pascal}
\vspace{-.3cm}
\end{figure*}

\section{Generated Samples and Qualitatives}
Fig.~\ref{fig:generated} shows additional image-text generated examples used for the presented positive-augmented contrastive learning strategy. As it can be seen, both image and text generated samples are realistic and plausible and can be effectively used as an additional source of supervision.

\begin{figure*}[t]
\centering
\includegraphics[width=\linewidth]{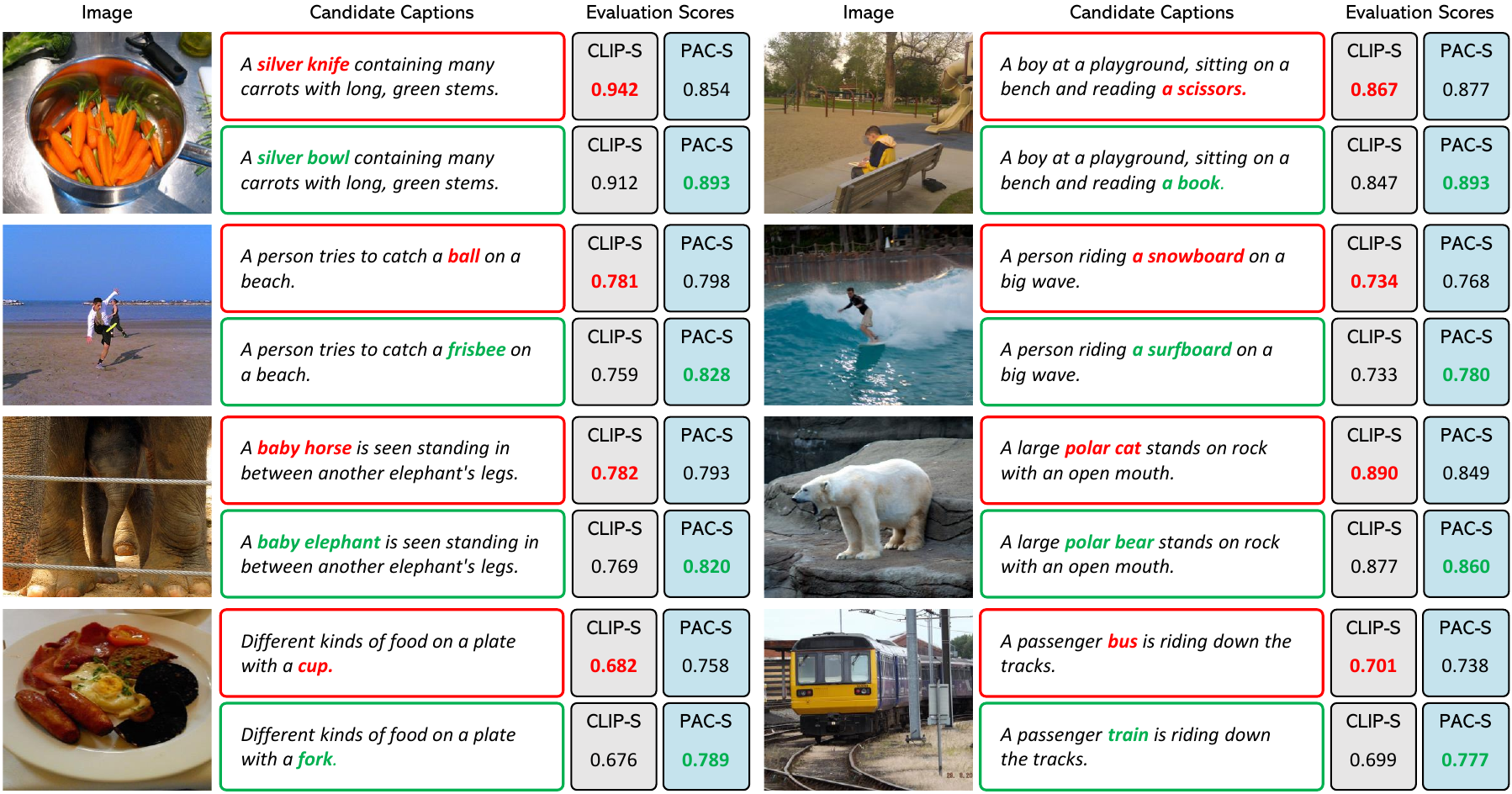}
\caption{Sample images from the FOIL hallucination detection dataset and corresponding evaluation scores generated by our proposed metric in comparison with CLIP-S. Captions with hallucinated objects are highlighted in red.}
\label{fig:foil}
\end{figure*}

\begin{figure*}[t]
\centering
\includegraphics[width=\linewidth]{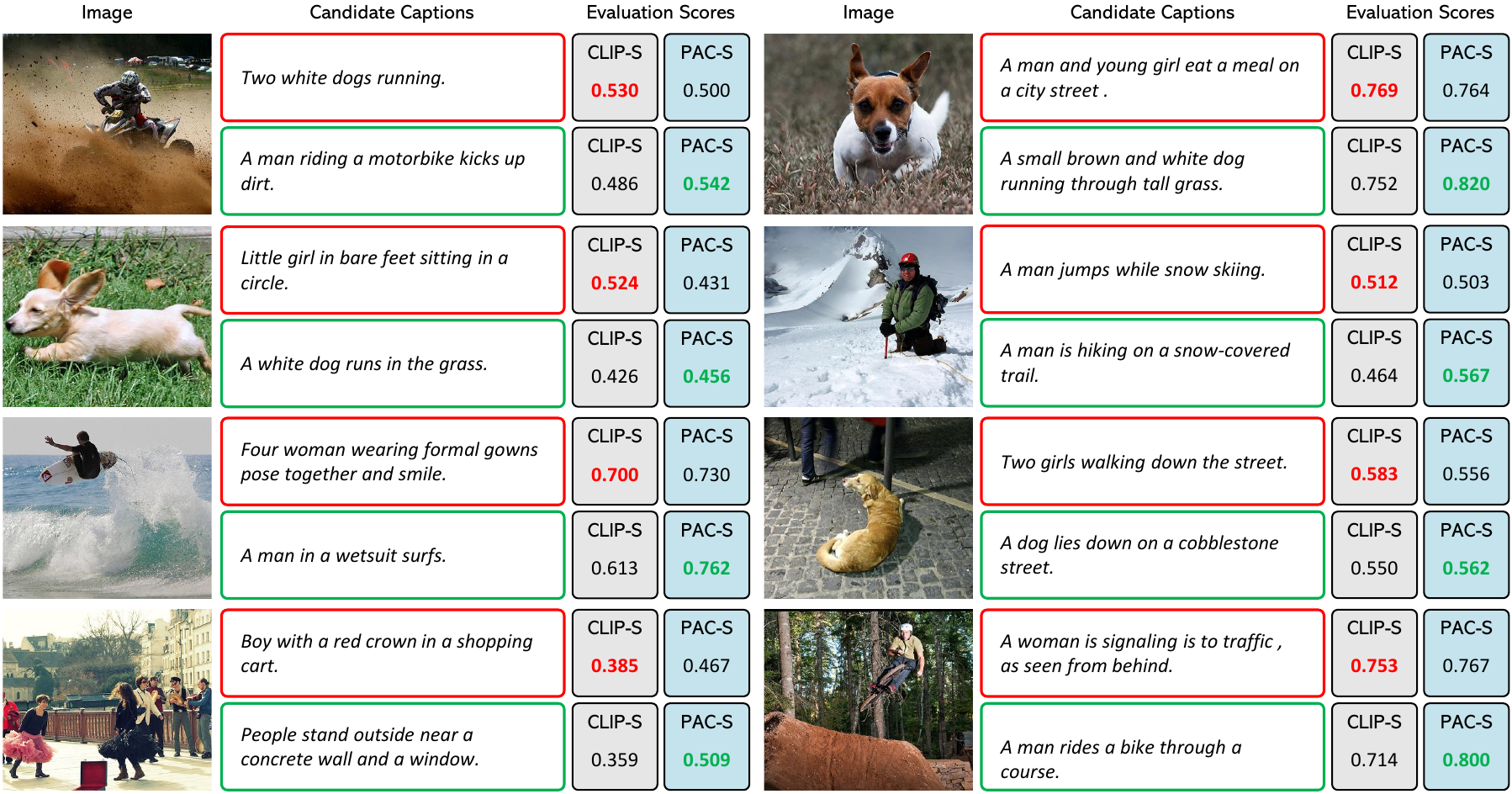}
\caption{Sample images from both Flickr8k-Expert and Flickr8k-CF datasets associated with the corresponding CLIP-S and \ours scores. The preferred caption accordingly to the human ratings is highlighted in green.}
\label{fig:flickr}
\end{figure*}

We report in Fig.~\ref{fig:pascal} some additional qualitative comparisons between \ours and well-known metrics on the Pascal-50S dataset. These qualitative results show that in the majority of cases \ours is more aligned with the human judgments than other metrics. Finally, in Fig.~\ref{fig:foil} and~\ref{fig:flickr}, we report sample results comparing our metric with CLIP-S~\cite{hessel2021clipscore} on FOIL, Flickr8k-Expert, and Flickr8k-CF datasets. As it can be observed, \ours can correctly identify hallucinated objects and better correlates with human judgments, demonstrating its effectiveness compared to CLIP-S also from a qualitative point of view.

